\UseRawInputEncoding
\pdfoutput=1
\documentclass[10pt,journal,compsoc]{IEEEtran}

\usepackage{multirow}
\usepackage{amsfonts}
\usepackage{rotating}
\usepackage{balance}
\usepackage{enumitem}
\usepackage{cite}
\usepackage{xcolor}
\usepackage{graphicx}
\usepackage{bbm}
\usepackage{subcaption,siunitx,booktabs,caption}
\usepackage[textsize=tiny]{todonotes}
\usepackage{amsmath,amsthm}
\usepackage{float}
\usepackage{hyperref}
\newcommand*{\Scale}[2][4]{\scalebox{#1}{$#2$}}

\theoremstyle{plain}
\newtheorem{theorem}{Theorem}[section]

\newtheorem{example}[theorem]{Example}
\newtheorem{assumption}[theorem]{Assumption}
\theoremstyle{definition}
\newtheorem{definition}[theorem]{Definition}
\theoremstyle{remark}
\newtheorem{remark}[theorem]{Remark}

\definecolor{Red}{rgb}{1,1,1}
\definecolor{Blue}{rgb}{0,0,1}

\begin{document}
\title{Instrumental Variables in Causal Inference and Machine Learning: A Survey}

\author{Anpeng~Wu, Kun~Kuang, Ruoxuan~Xiong, Fei~Wu,~\IEEEmembership{Senior~Member,~IEEE}

\IEEEcompsocitemizethanks{
    \IEEEcompsocthanksitem 
    A. Wu, K. Kuang and F. Wu are with the College of Computer Science and Technology, Zhejiang University, China.
    \protect
    (E-mail: anpwu@zju.edu.cn; kunkuang@zju.edu.cn; wufei@cs.zju.edu.cn).
    \IEEEcompsocthanksitem R. Xiong is with the Department of Quantitative Theory \& Methods, Emory University, USA.
    \protect
    (E-mail: ruoxuan.xiong@emory.edu).
    \IEEEcompsocthanksitem K. Kuang is the corresponding author.
}
}

\markboth{Journal of \LaTeX\ Class Files,~Vol.~14, No.~8, August~2015}%
{Shell \MakeLowercase{\textit{et al.}}: Bare Demo of IEEEtran.cls for Computer Society Journals}

\IEEEtitleabstractindextext{
\begin{abstract}
Causal inference is the process of using assumptions, study designs, and estimation strategies to draw conclusions about the causal relationships between variables based on data. This allows researchers to better understand the underlying mechanisms at work in complex systems and make more informed decisions.
In many settings, we may not fully observe all the confounders that affect both the treatment and outcome variables, complicating the estimation of causal effects. To address this problem, a growing literature in both causal inference and machine learning proposes to use Instrumental Variables (IV).
This paper serves as the first effort to systematically and comprehensively introduce and discuss the IV methods and their applications in both causal inference and machine learning. First, we provide the formal definition of IVs and discuss the identification problem of IV regression methods under different assumptions. Second, we categorize the existing work on IV methods into three streams according to the focus on the proposed methods, including two-stage least squares with IVs, control function with IVs, and evaluation of IVs. For each stream, we present both the classical causal inference methods, and recent developments in the machine learning literature. 
Then, we introduce a variety of applications of IV methods in real-world scenarios and provide a summary of the available datasets and algorithms. Finally, we summarize the literature, discuss the open problems and suggest promising future research directions for IV methods and their applications. We also develop a toolkit of IVs methods reviewed in this survey at \url{https://github.com/causal-machine-learning-lab/mliv}.  
\end{abstract}

}

\maketitle
\IEEEdisplaynontitleabstractindextext
\IEEEpeerreviewmaketitle

\IEEEraisesectionheading{
\section{Introduction}
\label{sec:introduction}
}

Nowadays, traditional machine learning and statistical modeling explore correlation patterns among observational variables for data mining and explanatory analysis, and have made amazing achievements in many domains over the past year \cite{wu2019intelligent, wu2020graph, lee2022deep}, especially in speech recognition, image recognition, natural language processing and recommender systems. 
As correlation-based algorithms, machine learning techniques gain striking performance from the over-fitting in training distributions under the IID hypothesis that training and testing data are independently sampled from the identical distribution.  
However, these models will degrade performance when the test distribution undergoes uncontrolled and unknown distribution shifts \cite{he2021towards, scholkopf2021towards}, i.e., Out of Distribution (OOD) setting. 
Essentially, the accuracy drop of current models is mainly caused by the spurious correlation between the features and labels \cite{shen2020stable, zhang2021deep}, refered as confounding bias\footnote{As introduced in Chapter 3.3 in Causality \cite{pearl2009causality}, the \textbf{confounding bias} between the feature input and target output can be defined as the bias of causal effect estimation when imbalanced confounders exist. Confounders are common causes of feature and ouput of interest. }. 
For example, if we do not consider the peak season, we may mistakenly conclude that higher airline ticket prices will lead to higher sales, as the peak season will lead to changes in both prices and demand for airline tickets. 
Lack of interpretability, actionability and stability from causality, correlation-based models has poor generalization performance on OOD data \cite{scholkopf2021towards,cui_stable_2022}. 


To address these issues, machine learning community has tried to develop causality-inspired models by incorporating causal inference paradigms. 
The substantive content of these paradigms is to exploit the invariant causal relationships in the data to build models and establish stable and interpretable predictions. Scholkopf and Bengio (2022) \cite{scholkopf2021towards} collectively refer to these approaches as structural causal models to answer counterfactual questions and make the model imaginative, that is, models can give a correct prediction in unseen scenarios.
To identify the stable causal effects rather than unstable correlation patterns, the gold standard approach is to perform Randomized Controlled Trials (RCTs), where different treatments are randomly assigned to units. Nevertheless, RCTs are unrealistic in some settings due to ethical and cost issues. 
Hence, various methods are developed to draw inference of causal effects from observational datasets, commonly under the unconfoundedness assumption, e.g., propensity score\cite{bang2005doubly,wang2019doubly}, covariate balance\cite{athey2018approximate,zubizarreta2015stable,hainmueller2012entropy}, back-door criteria \cite{pearl1995causal, pearl2009causality} and representation learning\cite{shalit2017estimating}. 
However, in practice, regardless of the approach that one adopts to control confounding in observational studies, there always exists the possibility of bias, when unmeasured confounders exist. 

\begin{figure}[t] 
\begin{center}
\includegraphics[width=0.8\linewidth]{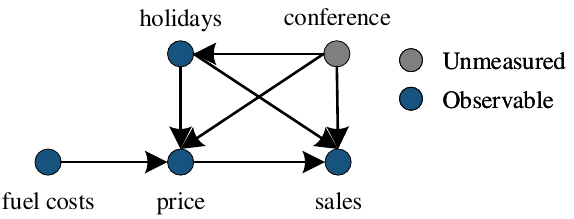}
\caption{The airline demand example.}\label{fig:8}
\end{center} 
\vspace{-0.1in}
\end{figure}

To control for unmeasured confounding, we introduce a third variable, named instrumental variable (IV), which is a cause of input features, has no direct effect on the outcome and does not share common causes with outcome. Using an instrumental variable to identify the hidden (unmeasured) correlation allows one to see the true correlation between the explanatory variable and response variable. For instance, the cost of fuel was used as an instrument in \cite{hartford2017deepiv} to estimate the impact of ticket prices on sales. Thus, changes in the cost of fuel create movement in ticket prices that is independent of unmeasured confounders, and this movement is equivalent to randomization for the purposes of causal inference \cite{hartford2017deepiv}. See Fig. \ref{fig:8} for a graphical illustration of this example and of the general class of causal graphs that we consider.

Two commonly used estimators for using an instrumental variable to estimate treatment effects are the two stage least squares estimator (2SLS) and the control function estimator (CFN) \cite{imbens2007control, guo2016control}: 
(1) 2SLS identifies the probability distribution over the treatment conditioned on the IVs in the treatment regression stage, and regresses the outcome based on the conditional distribution of the treatment (obtained from the treatment stage) in the outcome regression stage.
Based on the adopted model, we divide 2SLS and its variants into three categories as vanilla 2SLS estimator for linear models, sieve estimator and machine learning estimator for non-linear models.
(2) CFN constructs the residual variables (called control functions) in the estimation of treatment from the treatment regression stage, and estimates the outcome from the observed treatment and the residual in the outcome regression stage. 
Based on the structural assumption, we divide CFN and its variants into two categories as linear estimator and non-linear estimator. 

In linearity setting, \cite{guo2016control} show that CFN estimator is a 2SLS estimator with an augmented set (i.e., control function for unmeasured confounders) from instrumental variables. If these augmented variables are valid, then the control function estimator, while less robust than two stage least squares, might be much more precise because it keeps the treatment variables in the second stage \cite{imbens2007control}. However, if the augmented variables are not valid, then CFN estimator may be inconsistent, which is common in non-linear models. 
Fortunately, with more flexible kernel methods and neural network functions, machine learning methods have developed conditional density estimators, mutual information estimators, representational equilibrium models, etc. that can learn automatically from the data, for IV regression. This relaxes the linearity assumption and allows us to explore more complex causal systems and data. 

One limitation is that these standard methods and variants of instrumental variable (IV) analysis require a pre-defined strong valid IV. These methods are reliable only when the pre-defined IV only affects the outcome through its strong association with the cause variable of interest, in practice, which is hardly satisfied due to the untestable exclusion association with outcome. 
Therefore, in addition to lagged values and prior knowledge \cite{axinn2001mass,bollen2004automating,bollen2019model}, 
researchers usually implement Randomized Controlled Trials (RCTs) to sample a random variable as IV to intervene the received treatments, called intention-to-treat variable, such as Oregon health insurance experiment \cite{finkelstein2012oregon} and effects of military service on lifetime earnings \cite{angrist1990lifetime}, which are too expensive to be universally available.
To save the human effort selecting pre-defined IVs, a growing number of machine learning methods have been proposed to summary existing IV candidates to generate a valid IV representations \cite{burgess2017review,burgess2013use,kuang2020ivy,hartford2021valid,yuan2022auto}. 

In this paper, we provide a comprehensive review of the instrumental variable methods under the potential outcome framework. We first introduce the background of the potential outcome framework and instrumental variable, including the basic definitions, corresponding assumptions, and the fundamental problems with their general solutions. To identify the causal effect from instrumental variable, then, we further summarize most of the identification conditions of instrumental variables. Combined with machine learning, subsequently, we introduce the two-stage least-squares 
method (2SLS) and the traditional control function method (CFN) to estimate the average treatment effects. To void human effort selecting pre-defined IVs, we also discuss how to use machine learning algorithms to synthesize a summary-IV to plug into IV-based methods. 
Then,  we provide the related experimental information, including the available datasets that are
commonly adopted in the experiments, and the open-source codes of the above methods.
We also develop a toolkit of IVs methods reviewed in this survey at
\url{https://github.com/causal-machine-learning-lab/mliv}.  

Machine learning methods provide more flexible network models and conditional moment constraint models, which promote the development of causal inference. 
Meanwhile, causal inference also contributes to the development of machine learning methods.
Recently, advent works \cite{scholkopf2021towards,zhang2021deep, tang2022invariant} have revealed the existence and pervasiveness of variant and invariant(stable) features in data-driven algorithms and pointed out that the unstable features can provoke unexpected estimation bias for predictions.
Lacking a causal perspective, machine learning algorithms are prone to exploit subtle statistical correlations present in the training distribution for predictions, which is effective when testing data and training data are independently sampled from identical distribution, i.e., IID hypothesis.  
In practice, however, unbalanced samples and attribute-wise imbalance are common across different scenarios\cite{tang2022invariant}, unlike high-quality experimental data. That means estimators tend to regard high-frequency features from the training as predictive features and view low-frequency stable features as noise, which is unstable in other distributions and even bring additional bias. Due to low-quality observational data and some key unmeasured factors, there still exists a lack of common consensus on underlying invariant features in data-driven algorithms, albeit comprehensive endeavors \cite{scholkopf2021towards}. 
Therefore, researchers proposed instrumental variable regression to develop causality-inspired models, and the real-world applications that the discussed methods have great potential to benefit are discussed, including the social networks, recommendation system, computer vision, genome project, and domain adaptation as the representative examples. 

To the best of our knowledge, this is the first paper that provides a comprehensive survey for instrumental variable methods under the potential outcome framework. There also exist several surveys that discuss the causal effect estimation methods under the unconfoundedness assumption, \cite{guo2020survey,yao2021survey} introduce. 
To summarize, our contributions of this survey are as follows:
\begin{itemize}
  \item \emph{Comprehensive review}. We provide a comprehensive survey for instrumental variable methods under the potential outcome framework, including identification conditions, two-stage regression methods and control function algorithms.
  \item \emph{General setting}. When we cannot access a valid instrumental variable directly, we survey a line of IV testing methods and IV synthesis methods.
  \item \emph{Abundant resources}. In this survey, we list the state-of-art methods, the benchmark data sets, open-source codes, and representative applications. 
  \item \emph{Reproducible}. We integrate the existing resources and codes, and provide a unified interface and parameters to facilitate reproduction. 
\end{itemize}

The rest of the paper is organized as follows. In section 2, we introduce the background of the instrumental variable, including the basic definitions, the assumptions, and the fundamental problems with their general solutions. In section 3, we elaborate the structural assumption for identification of causal effect in IV regression. For estimation, the 2SLS-based methods and CFN-based methods are presented in Section 4 \& 5. In Section 6, we list a series of literature about IV selection and IV synthesis. Afterward, we provide experimental guidelines in section 7, and the typical applications of causal in Section 8. Final, in Section 9, we conclude several IV-based open problems and future directions. 

\begin{figure*}[t!] 
\begin{center}
\includegraphics[width=0.70\linewidth]{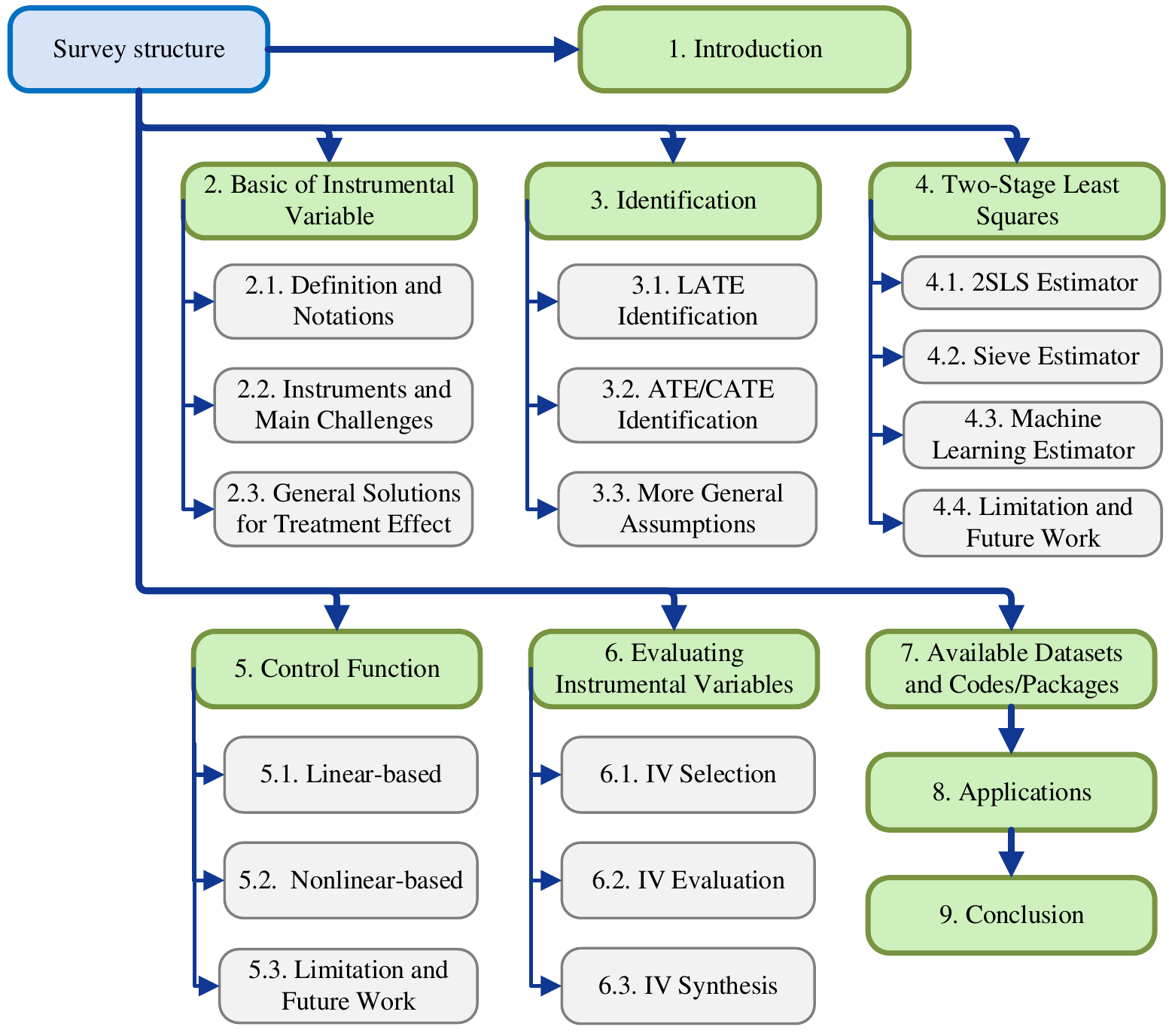}
\caption{Outline of the Survey.}\label{fig:overview}
\end{center} 
\vspace{-0.1in}
\end{figure*} 

\section{Basic of Instrumental Variable}
\label{sec:basic}

%
Although machine learning techniques have provided breakthroughs in statistics, econometrics, epidemiology and related disciplines, they usually suffer from low generalizability, instability, and inexplicability, due to the spurious relationship in which two or more events or variables are associated but not causally related\cite{guo2020survey,yao2021survey}. For example, in airplane sales (Example \ref{ex:1}), holidays and conferences may confound the causal relationship between prices and sales and introduce additional bias; in hospital (Example \ref{ex:2}), comorbidities and physical fitness would distort the causal relationships between the treatments and outcomes. Spurious relationship, deriving from confounders that are common causes of treatments and outcomes, is a common phenomenon in real-world scenarios. 
Hence, it is incredibly imperative and highly demanding to eliminate bias from conofunders and develop stable approaches to infer causal effect in observational studies, known as causal inference.

%
\begin{example}
\label{ex:1}
In the relationship (Fig. \ref{fig:8}) between airline ticket prices and sales, prices and demand rise and fall through the seasons, being affected by other events, such as holidays and conferences \cite{hartford2017deepiv, kato2021learning}. These events are called confounders that are the common causes of prices (cause variable, often referred to as the 'treatment') and sales (target outcome). 
\end{example}

%
\begin{example}
\label{ex:2}
In a hospital for infectious diseases, we study the effect of injection different from taking medicine (treatments) on patients' cure time (outcomes) from historical data. The patients' severity level of comorbidities and physical fitness are common causes of the treatments and outcomes, which we define as confounders. We may observe that patients with severe comorbidity have an injection, but the cure time is longer than those with mild comorbidity taking medicine, distorting the causal relationships between the treatments and outcomes.
\end{example}

%
In causal inference, the causal effects of treatment variable on target outcome, often referred to as the treatment effect, can be estimated using control experiments, regression models, matching estimators, re-weighting techniques, and instrumental variable (IV) \cite{yao2021survey, guo2020survey}. Among these approaches, the gold standard for treatment effect estimation is to perform Randomized Controlled Trials (RCTs), in which one of two or more treatments (cause variables) are randomly assigned to samples. With enough participants, RCTS would achieve sufficient control over confounding factors and deliver a useful comparison of the treatments studied. Considering the cost and ethical issues \cite{kohavi2011unexpected,bottou2013counterfactual}, fully RCTs are not always feasible in practical. Thus, in observational studies, there are a substantial number of regression models \cite{johansson2016learning,hassanpour2020learning}, matching estimators \cite{rosenbaum1983central,li2016matching}, and re-weighting techniques are developed to control or adjust the confounders to reduce the confounding bias under unconfoundedness assumption, i.e., all common causes of treatments and outcomes have been observed in data. 

%
Nevertheless, in real-world scenarios, it is common that unmeasured confounders exist, violating the unconfounderness assumption and posing a big challenge in estimating treatment effects from observational data. 
Regardless of the approach that one adopts to control confounding in observational studies, there always exists the possibility of bias due to unmeasured confounders \cite{brookhart2010confounding}, e.g., it is hard to obtain all conferences information in airline demand example (Fig. \ref{fig:8}). To overcome unmeasured confounder problems, researchers introduced an instrumental variable (fuel costs), an exogenous variable that induces changes in the treatments (prices) but has no independent effect on the outcomes (sales) \cite{wright1928tariff,hartford2017deepiv}, allowing researchers to uncover the causal effect of the treatment on the outcome under a series of identification assumptions developed by \cite{angrist1996identification,newey2003instrumental}. 

%
In 1928, the economist Philip Wright (Sewall's father) introduced IV, IV-estimator, and the equivalent two step least squares estimator, possibly in co-authorship with Sewall Wright, in the context of simultaneous equations in his book \emph{The Tariff on Animal and Vegetable Oils}\footnote{Based on \cite{stock2003retrospectives,hoveid2021constructing}.} \cite{wright1928tariff}. 
Later, Haavelmo \cite{haavelmo1943statistical} and Reiersøl \cite{reiersol1950identifiability} also applied the similar approach in the context of errors-in-variables models and contributed to the development of IVs unaware of the contributions of the Wrights. In linearity cases, IV estimators implement a two-stage least squares (2SLS) regression analysis for treatment effect estimation: stage 1 performs linear regression from the IVs to the treatments; and stage 2 performs linear regression from the conditional expectation of the treatments (obtained from stage 1) to the outcomes and the corresponding coefficient is used as a measure of treatment effect. To relax linearity assumption, \cite{pearl2000models,angrist1995identification,angrist1996identification} customized a series of identification assumptions for various scenarios, which would be elaborated in Section \ref{sec:identification}. 

%
The framework used by IV is essentially similar to potential outcome framework outlined by Rubin \cite{rubin1974estimating,rubin1978bayesian}. 
Next, we introduce the notations used in the IV estimator \cite{pearl2009causality}, and present the main challenges for causal effect estimation as well as general solutions for treatment effect.

\subsection{Definition and Notations}
\label{sec:notation}
\begin{figure}[H] 
\begin{center}
\includegraphics[width=0.72\linewidth]{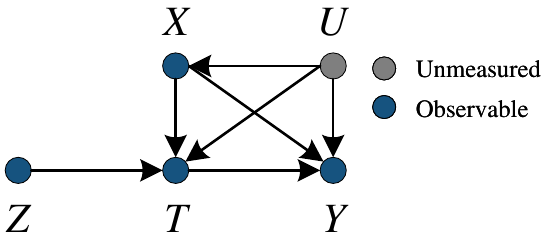}
\caption{The causal framework.}\label{fig:9}
\end{center} 
\vspace{-0.1in}
\end{figure}

%
The Rubin causal model \cite{rubin1974estimating,rubin1978bayesian,rubin1990comment}, also known as the potential outcome framework, is a standard approach for IV analysis and treatment effect estimation, named after Donald Rubin.
Similar to \cite{yao2021survey}, we define the notations under the potential outcome framework (Fig \ref{fig:9}). 

%
Note, in this paper, we use capital letters for random variables $(X)$, small letters for their values $(x)$, bold letters for vectors/sets of variables $(\mathbf{X})$ and their values $(\mathbf{x})$, and calligraphic letters for the spaces where they are defined $(\mathcal{X})$ if not explicitly stated. In addition, we use the subscript $i$ to represent the a variable $X_i$ belongs to the $i$-th unit, and view $x_i$ as specific value of $X_i$. 
To simplify notation, we consistently use the shorthand $p({x})$ to represent probabilities or densities $p({X}={x})$. For three random variables ${X}, {Y}, {Z}$, the conditional independence statement "${X}$ is conditionally independent of ${Y}$ given ${Z}={z}$" is written as ${X} \perp {Y} \mid {Z}$.

\begin{definition}
\textbf{Unit/Sample $i$}. A unit/sample denotes a single item or a collection of items from a larger whole or group. In the observational data, we can get a subset of $n$ samples from whole population, and we use the lowercase letter $i$ to mark each unit, $i=1,2,\cdots,n$. 
\end{definition}

\begin{definition}
\textbf{Treatment $T$}. Treatment refers to an intervention that applies (exposes, or subjects) to a unit. Based on the properties of treatments, we flesh out two cases: (1) in binary treatment cases, different treatment arms $T \in \{0,1\}$ denote different intervention (receive treatment or not) and researchers spilt all samples as the treated group ($T=1$) and the control group ($T=0$); (2) in multi-valued or continuous treatment cases, practitioners generalize the binary treatment effects framework and discrete or continuous interventions are used, called dose or dosage, i.e., $T \in \mathcal{T}, \mathcal{T} \subset \mathbb{R}$. 
\end{definition}

\begin{definition}
\textbf{Potential outcome $Y(T)$}. Potential outcome is a core element of potential outcome framework, which defines causal effect as a comparison between two states of the world, i.e., “factual” state and “counterfactual” state of the world. In the factual state, \textbf{the factual outcome} $Y(T=t)$ is \textbf{the observed outcome} of the treatment $T=t$ that is actually applied; in the counterfactual state, one would question "what would have happened if another treatment is applied" and imagine that same man takes another treatment $T=t'$ and get the \textbf{the counterfactual outcome} $\{Y(T=t')\}_{t' \not = t, t' \in \mathcal{T}}$. The above outcomes are called potential outcome $Y(T)$, which means a proposition stating what would have happened had a potential treatment $T$ been applied. 
\end{definition}

In the observational data, besides the treatment of interest and observed outcome, practitioners would collect other information for each units, which can be separated as pre-treatment variables and the post-treatment variables. 

\begin{definition}
\textbf{Pre-treatment variables} $\mathbf{V}=\{Z, X, U, \cdots \}$ are background variables that occur before the treatment $T$ is applied and will not be affected by the treatment $T$. 
Instead, a portion of the pre-treatment variables may be the causes of the treatment, and then researchers will assign treatment based on these variables for obtaining the desired outcome. 
\textbf{Post-treatment variables} $\mathbf{W}=\{Y, \cdots \}$ are variables that are affected by the
treatment $T$, and these events will occur after the treatment is accepted. 
In practice, based on the sequence of events and treatments occurring, Pre-treatment variables and Post-treatment variables are easily distinguished. 
In the following sections, we focus on the the pre-treatment variable $\mathbf{V}$ for causal inference, and we refer the terminology variable to the pre-treatment variable unless otherwise specified. 
\end{definition}

Both in decision-making applications and in the scientific literature, one tends to choose the level of the treatment to most efficiently pursue their objectives given the constraints they face \cite{imbens2014instrumental}. That means that the above pre-treatment variables may affect practitioners' treatment assignment, leading to unbalanced data distributions across different levels of the treatment. Recently, several works \cite{pearl2009causality,wu2022instrumental} shows such a unbalanced data would produce a spurious association, called confounding, because it tends to confound our judgment and to bias our estimate of the causal effect studied. For example, high-frequency but unrelated daily products are likely to be considered to exhibit correlation. 
Thus, \cite{pearl2009causality} claims that if a third variable $X$ that influences both $T$ and $Y$, the real and stable causal relationship may be confounded and spurious association would introduce additional bias for stable prediction. Such a variable is then called a confounder. 

\begin{definition}
\textbf{Confounders $\mathbf{X}$ \& $\mathbf{U}$}. 
In the causal relationship graph (Fig \ref{fig:9}), confounders are some special pre-treatment variables, which simultaneously affect the treatment assignment and the outcome being studied ($T,Y$) so that the effect estimation may not reflect the actual relationship ($T \rightarrow Y$) between the variables under study. 
In this paper, we denote $\mathbf{X}$ by \textbf{observable confounders} in observational data. For missing key variables in the record that may confound the relationship between the variables being studied ($T,Y$), we refer to them as \textbf{unmeasured confounders $\mathbf{U}$}. Confounders $\mathbf{X}$ \& $\mathbf{U}$ are both pre-treatment variables $\mathbf{V}$. 
\end{definition}

\noindent \textbf{Causal Inference}. After introducing the key terminologies and definition for causal inference, the causal effect can be quantitatively defined using the above definitions. 
In observational dataset, the treatment can be either binary, multi-valued or continuous.
For notational simplicity, we uniformly use $Y(T=t)$ to represent the potential outcome with treatment $T=t$. Then, the definition of the treatment effect is the difference $Y(T=t)-Y(T=0)$,  which can be measured at the population, subgroup, and individual levels. 

\begin{definition}
\textbf{Average Treatment Effect (ATE)}. 
\begin{eqnarray}
\textbf{ATE}(t) = \mathbb{E}[Y(T=t) - Y(T=0)],
\end{eqnarray}
\end{definition}

\begin{definition}
\textbf{Conditional Average Treatment Effect (CATE)}. 
\begin{eqnarray}
\textbf{CATE}(t,\mathbf{x}) = \mathbb{E}[Y(T=t) - Y(T=0) \mid \mathbf{X} = \mathbf{x}],
\end{eqnarray}
which has an another name: \\
\textbf{Individual Treatment Effect (ITE)}. 
\begin{eqnarray}
\textbf{ITE}_i(t) = Y_i(T=t) - Y_i(T=0).
\end{eqnarray}
\end{definition}

\subsection{Instruments and Main Challenges}
\label{sec:challenge}

In many circumstances, running Randomized Controlled Trials (RCTs) are not possible due to ethical or cost concerns. 
In the presence of unmeasured confounders $\mathbf{U}$, estimating treatment effect from observational data is challenging due to following reasons: 
\begin{itemize}
  \item \textbf{Counterfactual}. We only realize the outcome $y_i(T=t_i)$ with a specific treatment value $t_i$ applied to individual $i$, but cannot obtain the counterfactual outcomes $y_i(T \not = t_i)$ that would potentially happened if a different treatment option was assigned. 
  \item \textbf{Imbalanced observed Covariates}. The treatments are typically not assigned at random and the covariate distributions can be quite different between different treatment arms. Some high-frequency but unrelated variables $\mathbf{X}$ would confound the causal effect of treatment on outcome of interest. 
  \item \textbf{Imbalanced Unmeasured Covariates}. Even if we control all observed variables and adjust confounding differences from observational covariates, unmeasured key variables and differences $\mathbf{U}$ may distort the causal relationships in the observational data.
\end{itemize}

Hence, to overcome unmeasured confounder problems in observational data where the treatments are non-random assigned, researchers introduced an instrumental variable, an exogenous variable that induces changes in the treatments but has no direct effect on the outcomes, to estimate treatment effect. The instrumental variable is defined as follows:

\begin{definition}
\textbf{Instrument Variable $Z$} is an exogenous variable that affects the treatment $T$, but does not directly affect the outcome $Y$, as shown in Fig \ref{fig:9}. Besides, an valid instrument variable satisfies the following three restrictions: \\
\textbf{Relevance:} $Z$ is a cause of $T$, i.e., $\mathbb{P}(T \mid Z) \neq \mathbb{P}(T)$. \\
\textbf{Exclusion:} $Z$ does not directly affect the outcome $Y$, i.e., $Z \perp Y \mid T, \mathbf{X}, \mathbf{U}$.  \\
\textbf{Independent:} $Z$ is independent of all confounders, including $\mathbf{X}$ and $\mathbf{U}$, i.e., $Z \perp \mathbf{X},\mathbf{U}$
\end{definition}

Nevertheless, IV methods are reliable when the pre-defined IV is a valid IV that only affects the outcome through its strong association with treatment options, called exclusion assumption. Besides, they also need some strong structural assumptions, e.g., linear models. To sum up, IV regression has the following main challenges: 
\begin{itemize}
    \item \textbf{Strict Structural Assumption}. Even if the instrument $Z$ satisfies three restrictions in the definition, at least one structural assumption is required to identify the treatment effect of $T$ on $Y$ \cite{angrist1995identification,angrist1996identification,newey2003instrumental}. The most common structural assumption is the linearity assumption, which requires that the causal relationships between all variables are linear. 
    \item \textbf{Untestable Exclusion and Independent}. 
    We do not have access to the unmeasured confounders in observational data, and therefore we cannot test for independence between instrumental and unmeasured variables. In addition, we cannot test whether instrumental variables have additional causality on the outcome variable. 
    \item \textbf{Invalid and Weak IV}. In instrumental variables regression, the instruments are called \emph{weak IV} if their correlation with the endogenous regressors is close to zero, or \emph{invalid IV} if there is a direct effect or a hidden common cause between the instrument and the outcome. Due to untestable exclusion and independent restrictions, the predefined hand-made IVs could be weak or erroneous by violating the conditions of valid IVs.
\end{itemize}
Although IV has been used in tons of empirical papers, these thorny facts hinder the further application of the IV-based methods for treatment effect estimation. Recently, several works devote to relax or resolving these restrictions. 

For structural assumptions, a substantial number of IV works have been developed to relax the unconfoundedness assumption and the identification assumption for various scenarios \cite{heckman1990varieties,angrist1991sources,angrist1995identification,newey2003instrumental,newey2013nonparametric, hernan2020instruments,hartwig2020average}, which would be elaborated in Section \ref{sec:identification}. 
Although Exclusion and Independent are not testable, thanks to machine learning algorithms, researchers have developed Summary IV methods to automatically synthesize valid strong instrumental variables from a candidate set of instrumental variables \cite{burgess2013use,mokry2015mendelian,yuan2022auto}. Unless otherwise stated, in the following, we assume that the instrumental variables obtained from the observational data are valid strong IVs. 

\begin{remark} \label{remark:complierhhh}
Angrist, Imbens and Rubin \cite{angrist1995identification, angrist1996identification} abandoned the effort to draw inference for the overall average effect, and focused on sub-populations for which the average effect could be identified, the so-called compliers. In binary cases, where instrument variables ($Z$) are different intervention assignments and treatment variables are individuals' respond to assignments ($T(Z)$), four different compliance types defined by the pair of values ($T(Z=0), T(Z=1)$)\cite{imbens2014instrumental}:
\begin{eqnarray} \label{eq:complierhhh}
i \in \begin{cases}n \text { (never }-\text { taker }) & \text { if } T_i(0)=T_i(1)=0 \\ c \text { (complier }) & \text { if } T_i(0)=0, T_i(1)=1 \\ d(\text { defier }) & \text { if } T_i(0)=1, T_i(1)=0 \\ a \text { (always }-\text { taker }) & \text { if } T_i(0)=T_i(1)=1\end{cases}
\end{eqnarray}
The local average treatment effect or complier average causal effect is identified:
\begin{definition}
\textbf{Local Average Treatment Effect (LATE)}. 
\begin{eqnarray}
\textbf{LATE} = \mathbb{E}[Y_i(T=1) - Y_i(T=0) | i \in complier]
\end{eqnarray}
\end{definition}
\end{remark}

Under the monotonicity assumption \ref{assum:monotonicity}\footnote{Monotonicity Assumption would be elaborated in Section \ref{sec:identification}.},  the proportion of compliers can be obtained from the remainder:
\begin{eqnarray}
P(i \in c) = 1 - P(T=1|Z=0) - P(T=0|Z=1), 
\end{eqnarray}
Thus, monotonicity assumption is a sufficient identification assumption for LATE estimation. In this paper, we focus on reviewing more general identification assumption for ATE/CATE estimation (and thus also for LATE). 

\subsection{General Solutions for Treatment Effect}
\label{sec:general}

In IV Regression, there are two main frameworks for causal inference, i.e., Two-stage Least Squares (2SLS) \cite{hartford2017deepiv,singh2019kerneliv,muandet2019dualiv,lewis2020agmm} and Control Function Method (CFN) \cite{blundell2003endogeneity,petrin2010control, wooldridge2015control,puli2020general}. 
The former uses the conditional expectation of the treatment variable to estimate the causal effect, while the latter recovers the unmeasured confounders to estimate the causal effect.
In linearity assumption, we let $Z = [z_1,z_2,\cdots,z_n]^\prime$ and $T = [t_1,t_2,\cdots,t_n]^\prime$ and assume that the observational data is generated by:
\begin{eqnarray}
{T} = {Z} \alpha + \epsilon, {Y} = {T} \beta + \epsilon, 
\end{eqnarray}
where $\epsilon \sim \mathcal{N}(0,1)$, and $\{\alpha, \beta\}$ are the coefficients in the linear equation. Besides, our target is to predict the causal parameter $\beta$ as treatment effect estimation.
Details of the implementation of 2SLS and CFN are as follows. 

\subsubsection{Two-stage Least Squares (2SLS)}
2SLS identifies the probability distribution over the treatment conditioned on the IVs in the treatment regression stage, and regresses the outcome based on the conditional distribution of the treatment (obtained from the treatment stage) in the outcome regression stage. In linearity models, the predicted values from 2SLS are obtained: \\
Stage 1: Regress treatments ${T}$ on instruments ${Z}$:
\begin{eqnarray}
\hat{\alpha}= \arg\min_{\alpha}\sum_{i=1}^n(t_i - \alpha z_i)^2 = \left(Z^{\prime} Z\right)^{-1} Z^{\prime} T
\end{eqnarray}
Let $P_{Z}=Z\left(Z^{\prime} Z\right)^{-1} Z^{\prime}$, and the predicted treatment is:
\begin{eqnarray}
\widehat{T}=Z \hat{\alpha}=Z\left(Z^{\prime} Z\right)^{-1} Z^{\prime} T=P_{Z} T
\end{eqnarray}
Stage 2: Regress ${Y}$ on the predicted values $\widehat{T}$ from stage 1:
\begin{eqnarray}
\hat\beta = \arg\min_{\beta}\sum_{i=1}^n(y_i - \beta \hat{t}_i)^2
\end{eqnarray}
which gives:
\begin{eqnarray}
\beta_{\text{2SLS}}=\left(X^{\prime} P_{Z}^T P_{Z} T\right)^{-1} T^{\prime} P_{Z} Y
\end{eqnarray}
This method requires a strong linear relationship between the instrumental variables and the treatment variables, which will be not applicable if the unmeasured noise $\epsilon$ is large. More nonlinear variants of 2SLS are detailed in Section \ref{sec:2SLS}.

\subsubsection{Control Function Method (CFN)}
CFN constructs the residual variables (called control functions) in the estimation of treatment from the treatment regression stage, and estimates the outcome from the true treatment and the residual in the outcome regression stage. In linearity models, the predicted values from CFN are obtained: \\
Stage 1: Regress treatments ${T}$ on instruments ${Z}$:
\begin{eqnarray}
\hat{\alpha}= \arg\min_{\alpha}\sum_{i=1}^n(t_i - \alpha z_i)^2 = \left(Z^{\prime} Z\right)^{-1} Z^{\prime} T
\end{eqnarray}
and the predicted residuals is:
\begin{eqnarray}
\hat{\epsilon} = T - Z \hat{\alpha} = T - P_{Z} T
\end{eqnarray}
Stage 2: Regress ${Y}$ on the predicted residuals from stage 1:
\begin{eqnarray}
\hat\beta, \hat\beta_\epsilon = \arg\min_{\beta, \beta_\epsilon}\sum_{i=1}^n(y_i - \beta \hat{t}_i - \beta_\epsilon \hat{\epsilon}_i)^2
\end{eqnarray}
which gives:
\begin{eqnarray}
(\beta_{\text{CFN}}, \beta_{\epsilon})=
((T, \epsilon)^T(T, \epsilon))^{-1}(T, \epsilon)^T
Y
\end{eqnarray}
where $(A, B)$ means a concatenate of vectors/matrices $A$ and $B$. This method is valid for large unmeasured confounding bias. More nonlinear variants of CFN are detailed in the Section \ref{sec:cfn}.

\section{Identification}
\label{sec:identification}

In this section, we will discuss the structural assumptions or restrictions on data and model for precise inference to be possible, i.e., identification. After obtaining an infinite number of observations from population, if it is theoretically possible to learn the true values of a model's underlying parameters, then the model is identifiable. Otherwise, it is not non-identifiable. 
Even if the instrument satisfies IV's three constraints, we might not be able to identify causal effects unless there are additional structural assumptions \cite{chamberlain1986asymptotic,heckman1990varieties,angrist1991sources,angrist1995identification}. 

\begin{example}
\textbf{Non-identifiability} Without loss of generality, we take continuous treatment cases as an example and assume an unmeasured confound $U$ is a random variable from a standard normal distribution $\mathcal{N}(0,1)$: 
\begin{eqnarray}\label{eq:3}
T = Z U, Y = T U, U \sim \mathcal{N}(0,1).
\end{eqnarray}
Due to the unmeasured confounders, the relationships between $Z$ \& $T$, $Z$ \& $Y$ and $T$ \& $Y$ can no longer be accurately regressed by any parametric or non-parametric models. 
\end{example}

In econometric program evaluation, to address the non-identifiability problem, a standard method is to build structural equation model with linearity assumptions for treatment effect identification \cite{heckman1985alternative,heckman1989choosing}.

\begin{assumption} \textbf{Linearity Assumptions \cite{heckman1989choosing,angrist1996identification}. }
\label{assum:linear}
For experimental or observational data, let $Y$ be the observed outcome of interest, let $T$ be the observed treatment, and let $Z$ be the observed instrument variable. For continuous treatment cases, a standard structural assumption for the identification of treatment effect would have the form:
\begin{eqnarray}\label{eq:ass-linearCT}
T = \alpha_0 + \alpha_1 Z + \epsilon_T, \\
Y = \beta_0 + \beta_1 T + \epsilon_Y, \label{eq:ass-linearCY}
\end{eqnarray}
where $\{\alpha_0,\alpha_1,\beta_0,\beta_1\}$ are corresponding scalar coefficients, as well as $\epsilon_T$ and $\epsilon_Y$ are additive confounding effect from unmeasured confounders, which influence both the treatment and the outcome ($\epsilon_T \not \perp \epsilon_Y$). In the model $\beta_1$ represents the causal effect of $T$ on $Y$. \\
For binary treatment, the structural Eq. (\ref{eq:ass-linearCT}) could be reformulated as:
\begin{eqnarray}\label{eq:ass-linearBT}
T = 1\{ \alpha_0 + \alpha_1 Z + \epsilon_T \geq 0\}, 
\end{eqnarray}
where $1\{\cdot\}$ is a indicator function. 
\end{assumption}

Following IV's independent restriction, we have $Z \perp \epsilon_T, \epsilon_T$. Then, the absence of $Z$ in Eq. \ref{eq:ass-linearCY} denotes that any effect of $Z$ on $Y$ must be through an effect of $Z$ on $T$ in Eq. \ref{eq:ass-linearCT}/ \ref{eq:ass-linearBT}. Thus, $Z$ can be considered as a strong and valid IV for treatment effect estimation (i.e., $\beta_1$). The IV estimator is defined as 
the ratio of sample covariance \cite{durbin1954errors,angrist1996identification}. For binary instrument and treatment cases, 
\begin{eqnarray}\label{eq:linear-B}
\hat{\beta}_1 &=&\operatorname{cov}\left(Y, Z\right) / \operatorname{cov}\left(T, Z\right) \\ 
&=&
{\frac{\mathbb{E} (Y Z) / \mathbb{E} (Z)-\mathbb{E} (Y\left(1-Z\right)) / \mathbb{E} \left(1-Z\right)}{\mathbb{E} (T Z) / \mathbb{E} (Z)-\mathbb{E} (T\left(1-Z\right)) / \mathbb{E}\left(1-Z\right)} }
\end{eqnarray}
For continuous instrument and treatment cases, 
\begin{eqnarray}\label{eq:linear-C}
\hat{\beta}_1 &=&\operatorname{cov}\left(Y, Z\right) / \operatorname{cov}\left(T, Z\right) \\ 
&=& 
\frac{\mathbb{E} ( Y - (\mathbb{E} (Y))  ( Z - \mathbb{E} (Z)) )}{\mathbb{E} ( T - (\mathbb{E} (T))  ( Z - \mathbb{E} (Z)) )}
\end{eqnarray}

Without the linearity assumption in real-world scenarios, even if the instrument satisfies IV's three constraints, we might not be able to identify causal effects. 
However, the linearity assumption (Eq. \ref{eq:ass-linearCT}, \ref{eq:ass-linearCY} \& \ref{eq:ass-linearBT}) have 
not found widespread use in real-world scenarios, and exists only in theoretical studies. Besides, the apparently unreproducible experimental results also prevent the application of instrumental variable parameter models under the linear assumption \cite{lalonde1986evaluating}. 
To relax the linear assumption and avoid parametric evaluation models, researchers had devoted to establishing conditions that guarantee nonparametric identification of treatment effects in observational studies, i.e. identification without relying on functional form restrictions or distributional assumptions \cite{angrist1995identification, angrist1996identification, newey2003instrumental, chetverikov2017nonparametric, hartwig2020average}.

\noindent \textbf{Identifiability}. Briefly speaking, even if the IVs are valid, further assumptions are required for the identification of treatment effect. Following criticism of 
parametric evaluation models \cite{lalonde1986evaluating}, instead of sticking to the average treatment effects in a population of interest, researchers use some weaker assumptions to identify the average effect for the compliers sub-population, i.e., Local Average Treatment Effect (LATE). Sufficient assumptions for this include: Constant/Additive Treatment Effect \cite{chamberlain1986asymptotic}, Zero Probability on Some IV Value \cite{heckman1990varieties,angrist1991sources} and  Monotonicity \cite{angrist1995identification, angrist1996identification, chetverikov2017nonparametric}. 
However, under these assumptions, we can identify the average treatment effect for the group of compliers but not for the specific members. 

It was not until 2003, when Newey and Powell (2003) gave the identification and estimation results for nonparametric conditional moment restrictions, that practitioners started to focus on the identifiability of the structure function for outcome, i.e., Conditional Average Treatment Effect (CATE) or ATE \cite{newey2003instrumental,newey2013nonparametric,hartford2017deepiv}. Subsequently, some more general homogeneity assumptions are developed one after another \cite{hernan2006instruments,brookhart2007preference,wang2018bounded,hernan2020instruments,hartwig2020average} for ATE(CATE). 
In the econometrics literature, Homogeneity Assumption is a more general version than Monotonicity Assumption and Additive Noise Assumption \cite{hartwig2020average,hartwig2021homogeneity}. 

Based on assumption for LATE or CATE, we are going to elaborate on these assumptions as LATE Identification Assumptions, CATE Identification Assumptions, and More General Assumptions. 

\subsection{LATE Identification}
\label{sec:linearID}


As illustrated in Remark~\ref{remark:complierhhh}, Angrist, Imbens and Rubin \cite{angrist1995identification, angrist1996identification} abandoned the effort to draw inference for the overall average effect, and focused on sub-populations for which the average effect could be identified, the so-called compliers. Four different compliance types are defined in Eq.~\ref{eq:complierhhh}. The compliers means that the groups of people  who can be induced to change treatments by assigning different instruments. In this section, we will list some sufficient assumptions or conditions for identifying treatment effect of 
the compliers, as follows. 

\begin{assumption} \textbf{Constant Treatment Effect \cite{chamberlain1986asymptotic}. }
\label{assum:cte}
To prevent above problem, one condition is the treatment effect is constant: $\alpha = Y(1) - Y(0)$ for any unit $i$. Then $\mathbb{E}[Y \mid Z = z] - \mathbb{E}[Y \mid Z = w]$ is equal to:
\begin{eqnarray}\label{eq:9}
& & \mathbb{E}[Y \mid Z = z] - \mathbb{E}[Y \mid Z = w] \nonumber \\
& = & \mathbb{E}[T(z)Y(1) + (1-T(z)) \cdot Y(0) \mid Z = z] \nonumber \\
& - & \mathbb{E}[T(w)Y(1) + (1-T(w)) \cdot Y(0) \mid Z = w] \nonumber \\
& = & Y(1)P[T=1 \mid Z = z] + Y(0)P[T=0 \mid Z = z] \nonumber \\
& - & Y(1)P[T=1 \mid Z = w] + Y(0)P[T=0 \mid Z = w] \nonumber \\
& = & [Y(1)P(z) + Y(0)(1-P(z))] \nonumber \\
& - & [Y(1)P(w) + Y(0)(1-P(w))]  \nonumber \\
& = & (Y(1) - Y(0))(P(z) - P(w)) \nonumber \\
& = & \alpha (P(z) - P(w)) 
\end{eqnarray}
\end{assumption}

\begin{assumption} \textbf{Zero Probability \cite{heckman1990varieties,angrist1991sources}. }
\label{assum:zp}
A second approach is to assume the existence of some value of the instrument, $w \in \mathcal{W}$, such that the probability of participation conditional on that value is equal to zero, i.e., $P(w)=0$. Then $P(T(z)-T(w)=-1)=0$:
\begin{eqnarray}\label{eq:10}
& & \mathbb{E}[Y \mid Z = z] - \mathbb{E}[Y \mid Z = w] \nonumber \\
& = & \Scale[0.85]{ [Y(1)P(z) + Y(0)(1-P(z))] - [Y(1)P(w) + Y(0)(1-P(w))] } \nonumber \\
& = & \Scale[0.85]{ Y(1)P(z) + Y(0)(1-P(z)) - Y(0) } \nonumber \\
& = & \Scale[0.85]{ Y(1)P(z) - Y(0)P(z) } \nonumber \\
& = & \Scale[0.85]{ P(z) \mathbb{E}[Y(1)- Y(0) \mid T(z)=1] }
\end{eqnarray}
To identify the causal effect, we need to know at least one value $w$ of $\mathcal{W}$. 
\end{assumption}
Let $A$ be an indicator for the event $Z \not \in \mathcal{W}$, i.e., $A = \mathbbm{1}\{Z \not \in \mathcal{W}\}$. Then:
\begin{eqnarray}
& &\mathbb{E}[Y \mid A=0] \nonumber \\
& = & \mathbb{E}[Y \mid T(w)=1] \cdot P(w) \nonumber \\
& + & \mathbb{E}[Y \mid T(w)=0] \cdot (1-P(w)) \nonumber \\
& = & \mathbb{E}[Y(0)], w \in \mathcal{W}.
\end{eqnarray}
and 
\begin{eqnarray}
& &\mathbb{E}[Y \mid A=1] \nonumber \\
& = & \mathbb{E}[Y_0 \mid A=1] + P(z)\mathbb{E}[Y_1-Y_0 \mid T(z)=1] \nonumber \\
& = & \mathbb{E}[Y_0] + P(z)\mathbb{E}[Y_1-Y_0 \mid T(z)=1], z \not \in \mathcal{W}.
\end{eqnarray}
Since we can estimate $P(z)$, $\mathbb{E}[Y \mid A=0]$ and $\mathbb{E}[Y \mid A=1]$ and we know $z \not \in \mathcal{W}$, then we can identify ATE:
\begin{eqnarray}
\mathbb{E}[Y_1-Y_0 \mid T(z)=1] = \frac{\mathbb{E}[Y \mid A=1] - \mathbb{E}[Y \mid A=0]}{P(z)}. 
\end{eqnarray}

\begin{assumption} \label{assum:monotonicity}
\textbf{Monotonicity \cite{angrist1995identification, angrist1996identification, chetverikov2017nonparametric}. }
For all possible value of instrument, $z$ and $w$, either $T(z) \geq T(w)$ for any unit $i$, or $T(z) \leq T(w)$ for any unit $i$. Without loss of generality, the assumption is satisfied with $T(z) \geq T(w)$: 
\begin{eqnarray}\label{eq:11}
& & \mathbb{E}[Y \mid Z = z] - \mathbb{E}[Y \mid Z = w] \nonumber \\
& = & \Scale[1.0]{ (P(z)-P(w)) \cdot \mathbb{E}[Y(1)- Y(0) \mid T(z) - T(w)=1] } \nonumber \\
\end{eqnarray}
\end{assumption}



\subsection{ATE/CATE Identification}
\label{sec:nonlinearID}

For LATE models, assumption \ref{assum:cte}, \ref{assum:zp} or \ref{assum:monotonicity} are sufficient for identification \cite{chamberlain1986asymptotic,heckman1990varieties,angrist1991sources,angrist1995identification}. Besides, in a linear outcome process, where the outcome process is a sum of the causal effect and zero-mean noise, zero covariance between the instruments $Z$ and unmeasured disturbances (confounders) $\mathbf{U}$, suffices for identify the causal effect \cite{newey2003instrumental,newey2013nonparametric}. 
In a nonparametric IV (NPIV) model for CATE, the moment restrictions that unmeasured disturbances has conditional mean zero given instruments is a necessary restriction  for identification, i.e., $\mathbb{E}[\mathbf{U} \mid Z] = 0$. 

\begin{assumption}
\textbf{Additive Noise Assumption / Separability Assumption \cite{newey2003instrumental,hartford2017deepiv}. } In the parametric/nonparametric model (Eq. (\ref{eq:noise})), the identification/uniqueness of $\hat{g}(\mathbf{X},T)$ is equivalent to the nonexistence of any function $\delta(\mathbf{X},T): = g(\mathbf{X},T) - \hat{g}(\mathbf{X},T) \not = 0$ such that $\mathbb{E}[\delta(\mathbf{X},T) \mid Z] = 0$. 
\end{assumption}

In the nonparametric setting, the relationship between the outcome process and reduced form belongs a 1st Fredholm integral equation \cite{kress1989linear} and leads an ill-posed inverse problem \cite{newey2003instrumental}. Considering the identification of a general nonparametric model:
\begin{eqnarray}
Y = g(\mathbf{X}, T) + \mathbf{U}, \mathbb{E}[\mathbf{U} \mid Z] = \mathbb{E}[\mathbf{U}] = 0.
\label{eq:noise}
\end{eqnarray}
where $g(\cdot)$ denotes a true, unknown structural function of interest. For a consistency estimation, \cite{newey2003instrumental,newey2013nonparametric, hartford2017deepiv} identified the causal effect as the solution of an integral equation: 
\begin{eqnarray}
\mathbb{E}[Y \mid Z, \mathbf{X}] & = & \mathbb{E}[g(\mathbf{X}, T) \mid Z, \mathbf{X}] + \mathbb{E}[\mathbf{U} \mid \mathbf{X}] \nonumber \\
& = & \int \left[g(\mathbf{X},T) + \mathbb{E}[\mathbf{U} \mid \mathbf{X}]\right] dF(T \mid Z,\mathbf{X}) \nonumber \\
& = & \int \hat{g}(\mathbf{X},T) dF(T \mid Z,\mathbf{X})
\end{eqnarray}
where $F$ denotes the conditional cumulative distribution function of $T$ given $\{Z,\mathbf{X}\}$. Given two observable functions $\mathbb{E}[Y \mid Z, \mathbf{X}]$ and $F(T \mid Z,\mathbf{X})$, $\hat{g}(\mathbf{X},T)$ is the solution of the inverse problem. Then, we can identify ATE:
\begin{eqnarray}
\text{ATE} = \hat{g}(\mathbf{X},T) - \hat{g}(\mathbf{X},0) = g(\mathbf{X},T) - g(\mathbf{X},0). 
\end{eqnarray}
Therefore, \cite{newey2003instrumental,hartford2017deepiv} characterized identification of structural functions as completeness of certain conditional distributions $\mathbb{E}[\mathbf{U} \mid Z] = 0$. 

\subsection{More General Assumptions}
\label{sec:assumption}

In the econometrics literature \cite{hartwig2020average,hartwig2021homogeneity}, Homogeneity Assumption is a more general version than Monotonicity Assumption and Additive Noise Assumption.  Next, we describe two general Homogeneity Assumptions and the No Effect Modification Assumption. Note that the previous assumptions (except the Monotonicity Assumption) can be viewed as a special case of the Homogeneity Assumptions. 

\begin{assumption}
\label{assumZT}
\textbf{Homogeneous Instrument-Treatment Association \cite{brookhart2007preference,wang2018bounded,hartwig2020average}}: The association between the IV and the treatment is homogeneous in the different level of unmeasured confounders, i.e., $\mathbb{E}[T|Z=a,  \mathbf{U}]-\mathbb{E}[T|Z=b, \mathbf{U}]=\mathbb{E}[T|Z=a]-\mathbb{E}[T|Z=b]$.
\end{assumption}

\begin{assumption}
\label{assumTY}
\textbf{Homogeneous Treatment-Outcome Association \cite{hernan2006instruments,hernan2020instruments,hartwig2020average}}: The association between the treatment and the outcome is homogeneous in the different level of unmeasured confounders, i.e., $\mathbb{E}[Y|T=a,  \mathbf{U}]-\mathbb{E}[Y|T=b, \mathbf{U}]=\mathbb{E}[Y|T=a]-\mathbb{E}[Y|T=b]$.
\end{assumption}

Meanwhile, No effect modification of the treatment effect (NEM) is weaker than Homogeneity Assumptions, but may not be plausible in many instances \cite{hernan2020instruments,hartwig2020average}. 

\begin{assumption}
\label{assumNEM}
\textbf{No Effect Modification \cite{hernan2020instruments,wang2018bounded,hartwig2020average}}: The unmeasured confounders $\mathbf{U}$ would not modify the causal effect of $T$ on $Y$. 
\end{assumption}

\section{Two-Stage Least Squares}
\label{sec:2SLS}

\begin{figure*}[t!] 
\begin{center}
\includegraphics[width=0.92\linewidth]{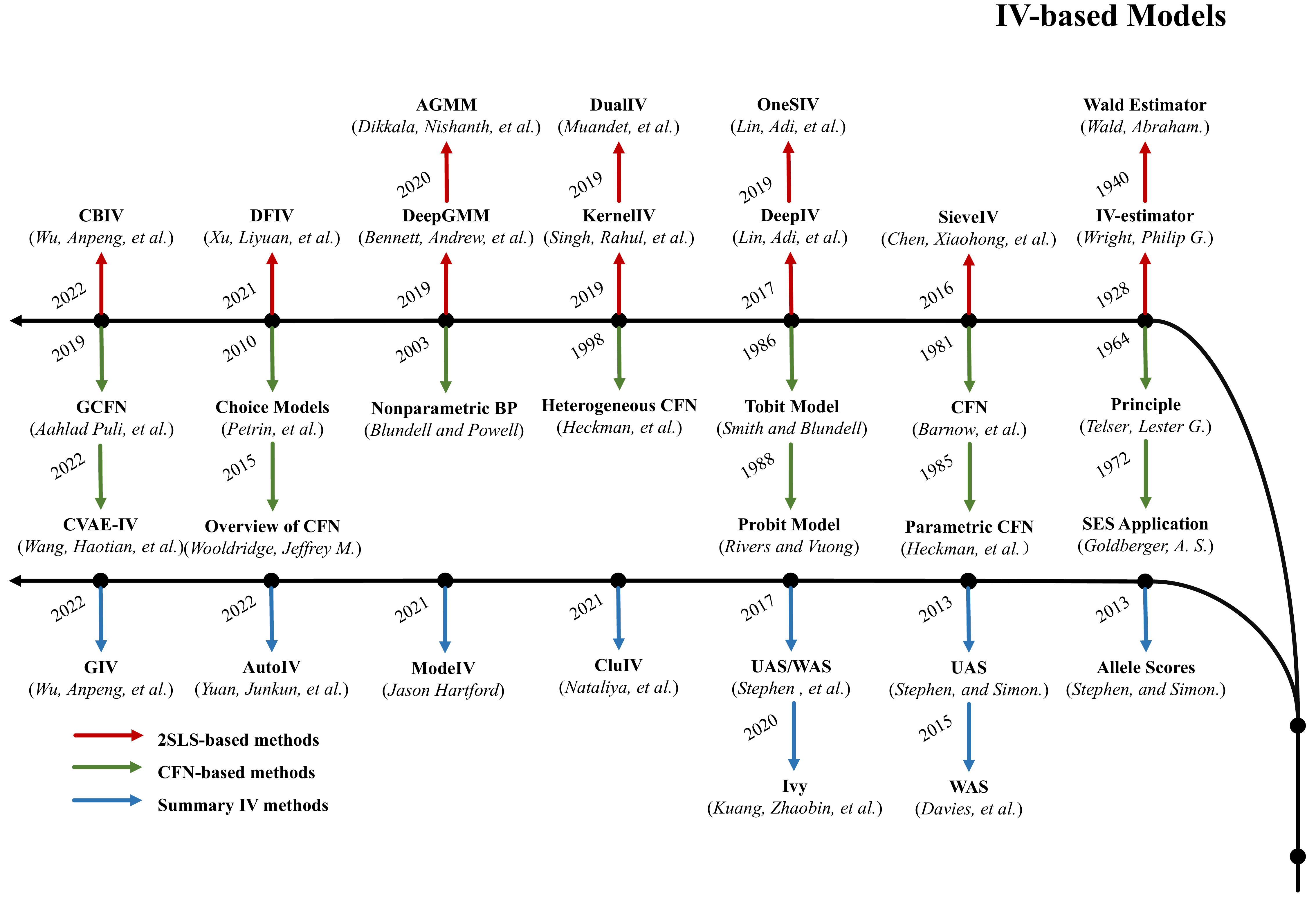}
\caption{Key milestones in the development of instrumental variable.}\label{fig:milestone}
\end{center} 
\vspace{-0.1in}
\end{figure*} 

As discussed above, we know that when there are unmeasured confounder in the data, the causality obtained by direct regression (Ordinary Least Squares, OLS) will be distorted. In this section, we will explain the inconsistency of ordinary least squares for causal effect and introduce some typical IV-based methods for consistency estimation, i.e., two-stage least squares and its variants with machine learning. 
As a classical statistical method for causal effect estimation, the two-stage least squares performs linear regression from the instruments $Z$ to the treatments $T$ in stage 1, and fit the counterfactual outcome function to predict the outcomes $Y$ from the conditional expectation of the treatments $\mathbb{E}[{T \mid Z}]$ (obtained from stage 1) in stage 2.

\begin{figure}
\begin{center}
\includegraphics[width=0.88\linewidth]{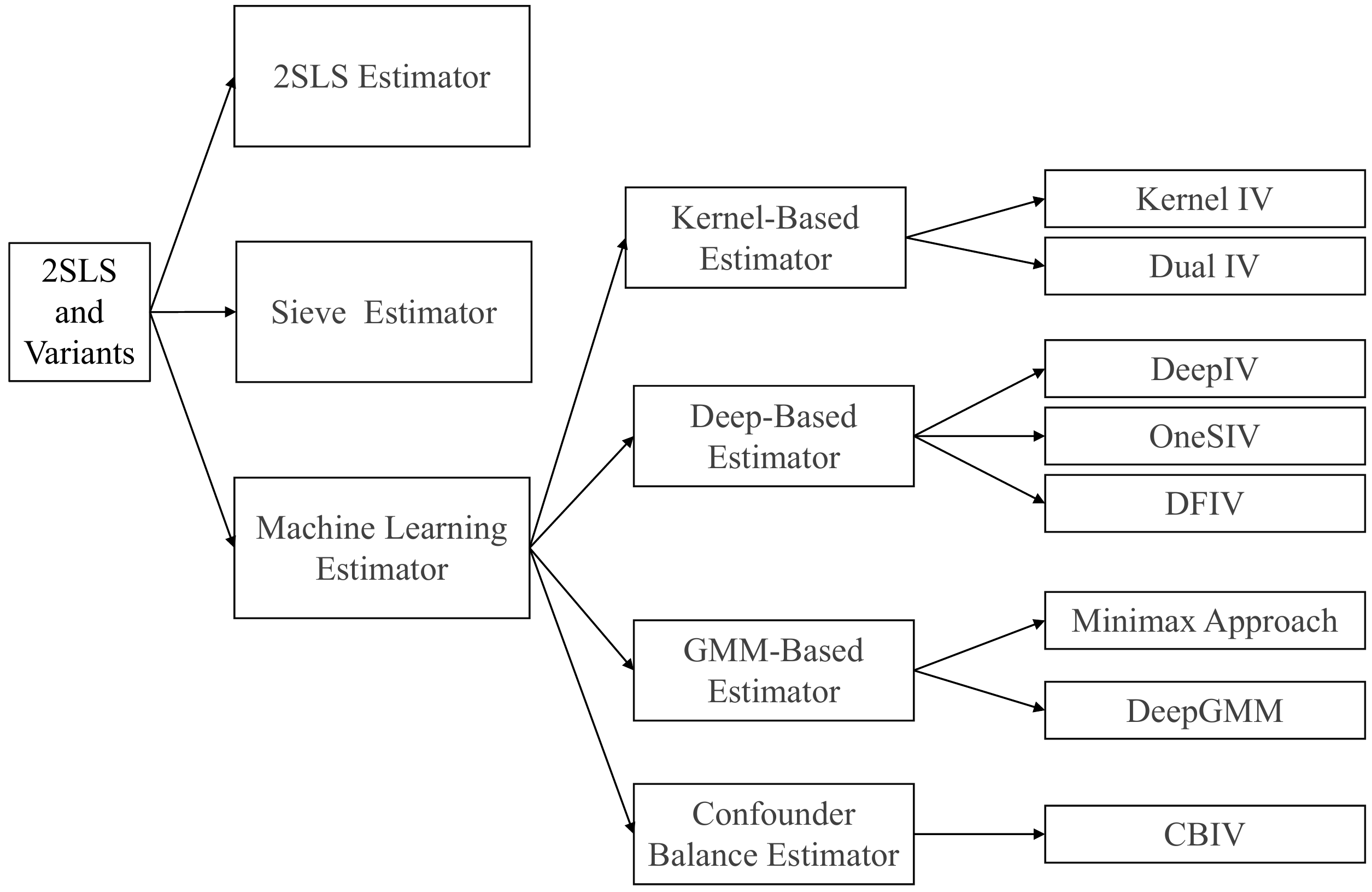}
\caption{Categorization of 2SLS and variants.}\label{fig:mliv}
\end{center} 
\end{figure}

Under the nonparametric identification of ATE/CATE in observational studies (See Section \ref{sec:nonlinearID}), based on traditional linear methods and advanced non-linear variants, as shown in Fig. \ref{fig:mliv}, we divide two-stage least squares and its variants into three categories: (1) Vanilla 2SLS and Wald Estimator for linear models (OLS is not applicable to causal effects); (2) Sieve estimator for non-linear models \cite{newey2003instrumental,chen2018optimal}; (3) Machine Learning for further estimation. There are four main research lines from machine learning estimator, incliuding: Kernel-based estimator \cite{singh2019kerneliv,muandet2019dualiv}, Deep-based estimator \cite{hartford2017deepiv,lin2019one,xu2021learning}, Moment conditions estimator \cite{lewis2020agmm, bennett2019deep} and confounder balance estimator.Finally we will summarize the limitations of these approaches and future works.

\begin{figure*}
\begin{center}
\includegraphics[width=0.82\linewidth]{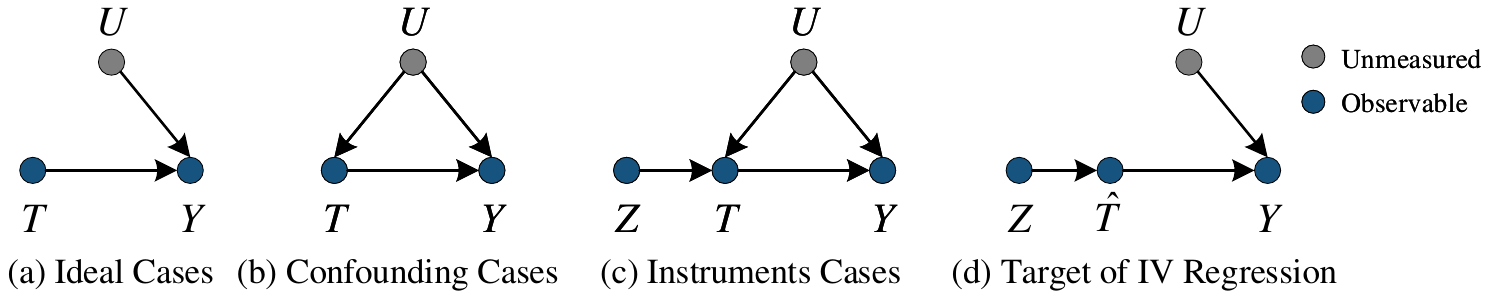}
\caption{The causal diagram in different cases.}\label{fig:diagram}
\end{center} 
\vspace{-0.1in}
\end{figure*} 

\subsection{2SLS Estimator}
\label{sec:wald}

Followed the linear Gaussian assumption in traditional 2SLS, without intercept for notational convenience, we assume that the observational data is generated by:
\begin{eqnarray}
\label{eq:linear_ZT}
T & = & Z \alpha + f(\mathbf{U})  =  Z \alpha + \epsilon_T, \\
\label{eq:linear_TY}
Y & = & T \beta + g(\mathbf{U})  =  T \beta + \epsilon_U, 
\end{eqnarray}
where $\{\alpha, \beta\}$ are the coefficients in the linear equation. Without interactions between unmeasured confounders and treatment, we can represent the effect of infinitely many unmeasured causes $\{f(\mathbf{U}),g(\mathbf{U})\}$ as an additive noise $\{\epsilon_{T}, \epsilon_Y\}$ regardless of how they interact among themselves, where $f(\cdot)$ and $g(\mathbf{\cdot})$ can be any continuous functions. Besides, the instrumental variable $Z$ is correlated with the independent variable $T$ and uncorrelated with the unmeasured confounder $U$. 
We call Eqs. (\ref{eq:linear_ZT}) and (\ref{eq:linear_TY}) the structural equations or primary equations, especially, Eq. (\ref{eq:linear_ZT}) is treatment-assignment function and Eq. (\ref{eq:linear_TY}) is counterfactual function in continuous setting. The corresponding causal diagram is shown in Fig. \ref{fig:diagram}(c).


\subsubsection{Inconsistency of Ordinary Least Squares}

In the presence of unmeasured confounder in the observational data, the causality obtained by direct regression (Ordinary Least Squares, OLS) will be distorted. 
In causal inference, the goal of regression analysis is to estimate the conditional expectation function $\mathbb{E}[Y \mid do(T)]$, i.e., to recover coefficient $\beta$ from the scalar regression model\footnote{$do(\cdot)$ denotes do-operation which manipulates the value of treatments as T in experimental study.}. Recall that ordinary least squares (OLS) solves for $\hat{\beta}$ by minimize the sum of squared errors:
\begin{eqnarray}
\label{eq:OLS}
\min _{\beta} \quad (Y-T \beta)^{\prime}(Y-T \beta).
\end{eqnarray}
The first-order condition is $T^{\prime}(Y-T \hat{\beta}) = T^{\prime} \hat{\epsilon}_Y = 0$. The regression results are reliable only when $T$ and $\mathbf{U}$ are independent $\mathbb{P}(T \mid \mathbf{U}) = \mathbb{P}(T)$, i.e., ${\epsilon}_T=f(\mathbf{U}) \equiv 0 $ in dose-assignment function Eq. (\ref{eq:linear_ZT}), as shown in Fig. \ref{fig:diagram}(a). Then the treatment variable $T$ affects the outcome variable $Y$ only through ${T} \beta$, and there is no association between $T$ and $\mathbf{U}$. 


But in real-world scenarios, there may exist some unmeasured confounders $\mathbf{U}$ that are the common causes of the treatment $T$ and the outcome $Y$, i.e., ${\epsilon}_T = f(\mathbf{U}) \not \equiv 0 $ in dose-assignment function Eq. (\ref{eq:linear_ZT}), as shown in Fig. \ref{fig:diagram}(b). Now there is an association between $T$ and $\mathbf{U}$, i.e., ${\epsilon}_T$. Then, the true model is believed to have $T^{\prime} {\mathbf{U}} \not = 0$ in the presence of unmeasured confounders $\mathbf{U}$. 


Based on non-zero $\mathbf{U}$-$T$ association ($f(\mathbf{U}) \not \equiv 0 $), from Eq. (\ref{eq:linear_TY}) there is a direct effect ($T \beta$) and an indirect effect via $\mathbf{U}$ affecting $T$ which in turn generates an additional false correlation term between $T$ and $Y$. If we directly perform OLS regression (Eq. (\ref{eq:OLS})) to estimate the causal effect, OLS will combine these two effects to give a bias result, i.e., $\hat{\beta}_\text{OLS} \not = \beta$. In this case, the coefficient on the treatment $T$ is given:
\begin{eqnarray}
\hat{\beta}_\text{OLS} & = & \left(T^{\prime} T\right)^{-1} T^{\prime} Y 
= \left(T^{\prime} T\right)^{-1} T^{\prime} (T \beta + {\epsilon}_Y) \nonumber \\
& = & \beta + \left(T^{\prime} T\right)^{-1} T^{\prime} {\epsilon}_Y  \\
\label{eq:derivative}
\hat{\beta}_\text{OLS} & = & \frac{dY}{dT} = \beta + \frac{d{\epsilon}_Y}{dT} 
\end{eqnarray}

Therefore, the OLS estimates the bias effect $\beta + {d{\epsilon}_Y}/{dT}$ rather than the true effect $\beta$. In a conclusion, the OLS estimator is biased and inconsistent for causal inference in the presence of unmeasured confoudners. Therefore, the researchers proposed a two-stage regression method to eliminate confounding bias ${d{\epsilon}_Y}/{dT}$ \cite{wright1928tariff}. 

\subsubsection{Two-Stage Least Squares}

In the linear Gaussian model (Eqs. (\ref{eq:linear_ZT}) and (\ref{eq:linear_TY})) discussed above, assumption \ref{assum:monotonicity} is automatically satisfied. Thus, we can identify the causal effect via 2SLS using IVs $Z$, which is not related to $\mathbf{U}$, i.e., $Z^{\prime} {\epsilon}_T =Z^{\prime} {\epsilon}_Y = 0$.  

\quad \\ \noindent \textbf{The Treatment Regression Stage:} in stage 1 of 2SLS, estimator regresses treatment $T$ from IVs $Z$:
\begin{eqnarray}
& \hat{\alpha} = \left(Z^{\prime} Z\right)^{-1} Z^{\prime} T 
= \left(Z^{\prime} Z\right)^{-1} Z^{\prime} (Z \alpha +  {\epsilon}_T ) = \alpha \\
& \hat{T} = \mathbb{E}[T \mid Z] =  Z \alpha
\end{eqnarray}
According to the IVs' unconfounded assumption, there is no association between $\hat{T}$ and $\mathbf{U}$. Hence, as shown in Fig. \ref{fig:diagram}(d) the first-order condition $\hat{T}^{\prime}(Y-T {\beta}) = \hat{T}^{\prime}  {\epsilon}_Y  = 0$ is satisfied. 


\quad \\ \noindent \textbf{The Outcome Regression Stage:} in stage 2 of 2SLS, estimator regresses the outcome $Y$ based on the conditional expectation of the treatment $\hat{T}$ (obtained from stage 1):
\begin{eqnarray}
& \hspace{-8pt} \hat{\beta}_\text{2SLS} = \left(\hat{T}^{\prime} \hat{T}\right)^{-1} \hat{T}^{\prime} Y 
= \left(\hat{T}^{\prime} \hat{T}\right)^{-1} \hat{T}^{\prime} (\hat{T} \beta + {\epsilon}_Y) = \beta \\
& \hat{Y} = \mathbb{E}[Y \mid \hat{T}] =  \hat{T} \beta
\end{eqnarray}
Then, we can get the counterfactual function by replacing $\hat{T}$ with $T$:
\begin{eqnarray}
\hat{Y} = \mathbb{E}[Y \mid do(T)] = T \beta.
\end{eqnarray}

\subsubsection{Wald Estimator}

In 1940s, the economist Wald proposed the wald estimator for a non-continuous IV case where the instruments $Z$ is a binary instrument \cite{wald1940fitting}. 
Denote the sub-sample averages of $Y$ and $T$ by $\Bar{Y}_1$ and $\Bar{T}_1$ when $Z = 1$ and by $\Bar{Y}_0$ and $\Bar{T}_0$ when $Z = 0$. Then, we can get the derivatives:
\begin{eqnarray}
\frac{dY}{dZ} = \Bar{Y}_1 -\Bar{Y}_0 \\
\frac{dT}{dZ} = \Bar{T}_1 -\Bar{T}_0
\end{eqnarray}
therefore, the causal effect is:
\begin{eqnarray}
\hat{\beta}_\text{Wald} = \frac{\Bar{Y}_1 -\Bar{Y}_0}{\Bar{T}_1 -\Bar{T}_0}.
\end{eqnarray}

Wald Estimator is a binary IV version of 2SLS, under linearity assumption. 

\subsection{Sieve Estimator}


\begin{figure}
\begin{center}
\includegraphics[width=0.72\linewidth]{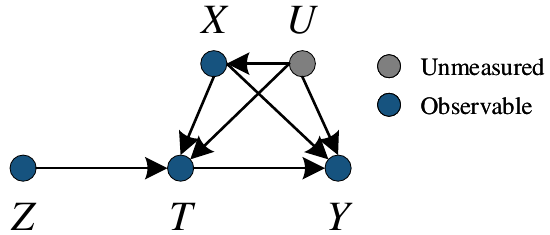}
\caption{The causal diagram in more general cases.}\label{fig:figureB}
\end{center} 
\vspace{-0.1in}
\end{figure} 

To satisfy the identification conditions, 2SLS simplifies the IV estimation problem by assuming linear models. To generalize 2SLS to the nonlinear setting, motivated by the works \cite{gallant1987identification,chen1998sieve} on sieve estimation, \cite{newey2003instrumental,newey2013nonparametric} propose a non-parametric two-stage basis expansion approach, called Sieve NPIV, with uniform convergence rates \cite{chen2018optimal}. Under a more general case ((Fig.~\ref{fig:figureB})), Sieve IV defines an appropriate finite dictionary of basis functions (Hermite polynomial or a set of indicator functions) for the treatments regression $T$ and the outcomes regression  $Y$ with instruments $Z$ and observed covariates $\mathbf{X}$, and specifies the number of basis expansion functions \cite{horowitz2011applied,chen2012estimation}. Specifically, under homogeneity assumption \ref{assumTY}, Sieve IV focus on identification of the models:
\begin{eqnarray}
\label{eq:fZX}
T = f(Z,\mathbf{X}) + \epsilon_T, \\
Y = g(T,\mathbf{X}) + \epsilon_Y, 
\label{eq:gTY}
\end{eqnarray}
where $g(\cdot)$ denotes the true, unknown structural function of interest, $\epsilon_T$ and $\epsilon_T$ are joint errors from unobserved variables $\mathbf{U}$, and the unmeasured confounders $\epsilon_T$ and $\epsilon_T$ are additive noise, that is independent with the instruments $Z$, i.e., $\mathbb{E}[\epsilon_T \mid Z] = \mathbb{E}[\epsilon_Y \mid Z] = 0$. 

Based on these sieve bases, Sieve IV implement a two-stage regression to estimate causal effect.

\quad \\ \noindent \textbf{Sieve IV}. \\
Formally, Sieve IV estimates the structure function using an appropriate finite dictionary of basis functions and we can reformulate the structure function as:
\begin{eqnarray}
\label{eq:34}
T = \sum_{i=1}^{d^Z}\sum_{j=1}^{d^X}\alpha_{i,j} \phi_i(Z) \xi_j(\mathbf{X}) + \epsilon_T,  \\
Y = \sum_{k=1}^{d^T}\sum_{j=1}^{d^X}\beta_{k,j} \psi_k(T) \xi_j(\mathbf{X}) + \epsilon_Y, 
\label{eq:35}
\end{eqnarray}
where $\{\phi_i\}_{i=1}^{d^Z}$ is the sieve basis for IVs $Z$ with degree $d^Z$, $\{\xi_i\}_{i=1}^{d^X}$ is the sieve basis for confounders $\mathbf{X}$ with degree $d^X$, $\{\phi_k\}_{k=1}^{d^T}$ is the sieve basis for treatments $T$ with degree $d^T$, and $\{\alpha_{i,j},\beta_{i,j}\}$ are the corresponding coefficients. Each of the $\phi_i$ is a function from $\mathcal{Z}$ into $\mathbb{R}$, each of the $\xi_j$ is a function from $\mathcal{X}$ into $\mathbb{R}$, and each of the $\psi_k$ is a function from $\mathcal{T}$ into $\mathbb{R}$. 

Then the goal of Sieve IV is to estimate:
\begin{eqnarray}
\text{CATE}(x,t) = \sum_{k=1}^{d^T}\sum_{j=1}^{d^X}\beta_{k,j} \xi_j(x) [\psi_k(T=t) - \psi_k(T=0)].
\end{eqnarray}

\noindent In \textbf{the treatment regression stage}, different than 2SLS, Sieve IV regresses each of the treatment basis functions ($\mathbb{E}[\psi_k(T) \mid \phi_i(Z)\xi_j(\mathbf{X})]$) on the basis features $\{\phi_i(Z)\xi_j(\mathbf{X})\}$ rather than the conditional expectation treatment distribution ($\mathbb{E}[T \mid Z, \mathbf{X}]$). 
\begin{eqnarray}
\hat{\alpha} = \text{argmin}_{\alpha} \text{MSE} (\psi(\sum_{i=1}^{d^Z}\sum_{j=1}^{d^X}[\alpha_{i,j} \phi_i(Z) \xi_j(\mathbf{X})]), \psi(T)).
\end{eqnarray}

\noindent In \textbf{the outcome regression stage}, Sieve IV estimates the expectation outcome onto these estimated functions $\mathbb{E}[\psi_k(T) \mid \phi_i(Z)\xi_j(\mathbf{X})]$ (obtained by the stage 1) and bases $\xi_j(\mathbf{X})$ to identify the coefficients $\beta_{k,j}$. 
\begin{eqnarray}
\hat{\beta} = \text{argmin}_{\beta} \text{MSE} (\sum_{k=1}^{d^T}\sum_{j=1}^{d^X}[\beta_{k,j} \psi_k(T) \xi_j(\mathbf{X})], Y).
\end{eqnarray}

In the two-stage regression of Sieve IV, the challenge is how to define an appropriate series basis functions \cite{chen2018optimal}. Thus, recent works \cite{singh2019kerneliv,muandet2019dualiv} introduce machine learning algorithm to obtain the basis functions and estimate causal effect. 

\subsection{Machine Learning Estimator}
To implement further estimation, as shown in Fig. \ref{fig:mliv}, there are four main research lines from machine learning estimator (Fig.~\ref{fig:mliv}), incliuding: Kernel-based Estimator \cite{chen2018optimal,singh2019kerneliv,muandet2019dualiv}, Deep-based methods \cite{hartford2017deepiv,lin2019one,xu2021learning}, Moment conditions methods \cite{lewis2020agmm, bennett2019deep} and Confounder Balanced Estimator \cite{wu2022instrumental}.

\subsubsection{Kernel-based Estimator}
\label{sec:kernel}

Motivated by Sieve NPIV \cite{chen2018optimal} and predictive state representation models (PSRs) \cite{boots2013hilbert} and \cite{hefny2015supervised}, \cite{singh2019kerneliv} proposes kernel instrumental variable regression (KernelIV) to model relations among $Z$, $\mathbf{X}$, $T$, and $Y$ as nonlinear functions in reproducing kernel Hilbert spaces (RKHSs) \cite{song2009hilbert}, and prove the consistency of KernelIV.

\begin{figure}
\begin{center}
\includegraphics[width=0.62\linewidth]{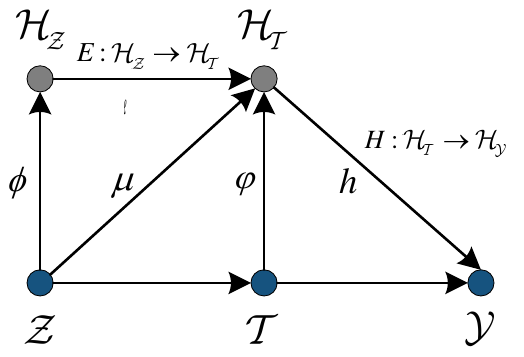}
\caption{The Structural Function of KernelIV.}\label{fig:kerneliv}
\end{center} 
\vspace{-0.1in}
\end{figure} 

\quad \\ \noindent \textbf{Kernel IV}. \\
As shown in Fig. \ref{fig:kerneliv}, KernelIV defines two measurable positive definite kernels $k_{\mathcal{T}}: \mathcal{T} \times \mathcal{T} \rightarrow \mathbb{R}$ and $k_{\mathcal{Z}}: \mathcal{Z} \times \mathcal{Z} \rightarrow \mathbb{R}$ corresponding to scalar-valued RKHSs $\mathcal{H}_{\mathcal{T}}$ and $\mathcal{H}_{\mathcal{Z}}$:
\begin{eqnarray}
\psi: \mathcal{T} \rightarrow \mathcal{H}_{\mathcal{T}}, t \mapsto k_{\mathcal{T}}(t, \cdot), \quad \phi: \mathcal{Z} \rightarrow \mathcal{H}_{\mathcal{Z}}, z \mapsto k_{\mathcal{Z}}(z, \cdot)
\end{eqnarray}
where $\psi$ and $\phi$ are the basis functions of $\mathcal{Z}$ and $\mathcal{T}$. 
In this section, $\mathcal{Z}$ means the the horizontal concatenation of IVs $\mathcal{Z}$ and confounders $\mathcal{X}$, and $\mathcal{T}$ means the the horizontal concatenation of treatments $\mathcal{T}$ and confounders $\mathcal{X}$, i.e., $\mathcal{Z} = \mathcal{Z} \oplus \mathcal{X}$ and $\mathcal{T} = \mathcal{T} \oplus \mathcal{X}$. Then, KernelIV reformulates the problem as:
\begin{eqnarray}
e \in E, E: \mathcal{H}_{\mathcal{Z}} \rightarrow \mathcal{H}_{\mathcal{T}}, \\
h \in H, H: \mathcal{H}_{\mathcal{T}} \rightarrow \mathcal{H}_{\mathcal{Y}}.
\end{eqnarray}

In stage 1, KernelIV learns a conditional mean embedding to model the relations between $\mathcal{Z}$ and $\mathcal{T}$ by two kernel functions $\psi$ and $\phi$ and a conditional expectation operator $E$:
\begin{eqnarray}
\hat{\phi}(T) = \mu({Z}) = e(\psi({Z})) = \mathbb{E}[\phi(T) \mid Z], \\
\psi(Z) \in {\mathcal{H}}_{\mathcal{Z}}, \quad \hat{\phi}(T), \phi(T) \in {\mathcal{H}}_{\mathcal{T}}, \nonumber
\end{eqnarray}
KernelIV constructs a objective for optimizing $e \in E$ by kernel ridge regression: 
\begin{eqnarray}
e_{\lambda}^* = \text{argmin}_{e \in E} \mathbb{E} \| [e\psi](Z) - \phi(T) \|^2 + \lambda  \| [e \|^2, \\
e_{\lambda}^* = \text{argmin}_{e \in E} \mathbb{E} \| \mu (Z) - \phi(T) \|^2 + \lambda  \| [e \|^2, 
\end{eqnarray}
where $\lambda$ is a hyper-parameter and $\| [e \|^2$ is a penalty term for function $e$. Indeed, $ \hat{T} = \mu({Z}) = [e^* \psi](Z)$. Analogously, in 2SLS $\hat{T} = \mathbb{E}[T \mid Z] = \hat{\alpha} Z$ for stage 1 linear regression parameter $\hat{\alpha}$.

In stage 2, to estimate the structural function $g(\cdot)$ (Eq. (\ref{eq:gTY})), KernelIV predicts the potential outcome function onto the conditional mean embedding $\hat{\phi}(T) \in {\mathcal{H}}_{\mathcal{T}}$:
\begin{eqnarray}
\hat{{Y}} = g({T}) = h ({\phi}(T)) = [h \mu]({Z}) = \mathbb{E}[{{Y}} \mid \hat{\phi}(T)], 
\end{eqnarray}
KernelIV constructs a objective for optimizing $h \in H$ by kernel ridge regression: 
\begin{eqnarray}
h_{\lambda}^* = \text{argmin}_{h \in H} \mathbb{E} \| h ({\phi}(T)) - Y \|^2 + \lambda  \| [h \|^2, \\
h_{\lambda}^* = \text{argmin}_{h \in H} \mathbb{E} \| h ({\mu}(Z)) - Y \|^2 + \lambda  \| [h \|^2, 
\end{eqnarray}
where $\| [h \|^2$ is a penalty term for function $h$.
Indeed, $ \hat{Y} = g({T}) = [h \phi]({T}) = [h \mu]({Z})$.  Analogously, in 2SLS $\hat{Y} = \mathbb{E}[Y \mid T] = \hat{\beta} T$ for stage 2 linear regression parameter $\hat{\beta}$. 

\quad \\ \noindent \textbf{Dual IV}. \\
Inspired by stochastic programming \cite{dai2017learning, shapiro2014lectures}, DualIV \cite{muandet2019dualiv} shows that two-stage IV-based regression can be reformulated as a convex-concave saddle-point problem. Then, \cite{muandet2019dualiv} develops a simple kernel-based algorithm and simplifies traditional two-stage methods via a dual formulation.

Based on the outcome structural function, the expectation of Eq. (\ref{eq:gTY}) w.r.t. $Y$ conditioned on $\{Z,\mathbf{X}\}$ yields \cite{newey2003instrumental}:
\begin{eqnarray}
\mathbb{E}[Y \mid Z,\mathbf{X}] & = & \mathbb{E}[g(T,\mathbf{X}) \mid Z,\mathbf{X}] + \mathbb{E}[{\epsilon}_Y \mid \mathbf{X}] \nonumber \\
& = &  \int g(T,\mathbf{X}) dF(T \mid Z,\mathbf{X}), 
\end{eqnarray}
where, $dF(T \mid Z,\mathbf{X})$ is the conditional treatment distribution obtained from the treatment regression. \cite{muandet2019dualiv} reformulate the equation as an empirical risk minimization problem: 
\begin{eqnarray}
\min_{g \in \mathcal{G}} R(g)=\mathbb{E}_{Y Z}\left[\ell\left(Y, \mathbb{E}_{T \mid Z,\mathbf{X}}[g(T,\mathbf{X})]\right)\right]
\label{eq:dualivERM}
\end{eqnarray}
where $\ell\left(y, y^{\prime}\right)=\left(y-y^{\prime}\right)^{2}$ denotes the mean squared error. 

Applying the interchangeability and Fenchel duality \cite{shapiro2014lectures,dai2017learning} to Eq. (\ref{eq:dualivERM}):
\begin{eqnarray}
R(g) = \mathbb{E}_{Y Z \mathbf{X}}\left[ \max _{u \in \mathbb{R}}\left\{\mathbb{E}_{T \mid Z,\mathbf{X}}[g(T,\mathbf{X})] u-\ell^{\star}(Y,u)\right\} \right] \nonumber \\
= \max_{u \in \mathcal{U}} \mathbb{E}_{Z \mathbf{X} Y}\left[\mathbb{E}_{T \mid Z, \mathbf{X}}[g(T,\mathbf{X})] u(Y, Z, \mathbf{X})-\ell^{\star}(Y,u(Y, Z, \mathbf{X}))\right] \nonumber \\
= \max_{u \in \mathcal{U}} \mathbb{E}_{Z \mathbf{X} T Y}[g(T,\mathbf{X}) u(Y, Z, \mathbf{X})]-\mathbb{E}_{Z \mathbf{X} Y}\left[\ell^{\star}(Y,u(Y, Z, \mathbf{X}))\right]  \nonumber
\end{eqnarray}
where $\mathcal{U}(\Omega)=\{u(\cdot): \Omega \rightarrow \mathbb{R}\}$ is the entire space of functions defined on the support $\Omega$, and $\Omega$ is the corresponding space of random variables $\mathcal{Y} \oplus \mathcal{Z} \oplus \mathcal{X}$. $\ell: \mathbb{R} \times \mathbb{R} \rightarrow \mathbb{R}_{+}$ is a proper, convex, and lower semi-continuous loss function for any value in its first argument and $\ell_{y}^{\star}=\ell^{\star}(y, \cdot)$ is a convex conjugate of $\ell_{y}=\ell(y, \cdot)$. 


To simplify notation, in this section, we denotes by $W=Y \oplus Z \oplus \mathbf{X}$ and $T = T \oplus \mathbf{X}$. Then, the saddle-point problem is:
\begin{eqnarray}
\min_{g \in \mathcal{G}} \max_{u \in \mathcal{U}} \mathbb{E}_{T W}[g(T) u(W)]-\mathbb{E}_{W}\left[\ell^{\star}(Y,u(W))\right] 
\end{eqnarray}
With $\ell^{\star}(y,u)=u y+\frac{1}{2} u^{2}$, DualIV reduce the traditional two-stage methods as:
\begin{eqnarray}
& \min_{g \in \mathcal{G}} \max_{u \in \mathcal{U}} \Psi(g, u), \\
& \hspace{-18pt} \Scale[1.0]{\Psi(g, u) = \mathbb{E}_{TW}\{[g(T) - Y] u(W)\} - \frac{1}{2} \mathbb{E}_{W}[u(W)^2]}.
\end{eqnarray}

Motivated by the reproducing kernel Hilbert spaces (RKHSs)\cite{song2009hilbert}, DualIV introduces positive definite kernels $k: \mathcal{T} \times \mathcal{T} \rightarrow \mathbb{R}$ and $l: \mathcal{W} \times \mathcal{W} \rightarrow \mathbb{R}$ for $\mathcal{G}$ and $\mathcal{U}$, respectively. \cite{scholkopf2002learning} introduces the canonical feature maps:
\begin{eqnarray}
\phi: t \mapsto k(t, \cdot), \varphi: w \mapsto l(w, \cdot). 
\end{eqnarray}

The objective can be rewritten as:
\begin{eqnarray}
\Psi(f, u) & = & \mathbb{E}_{T W}[f(T) u(W)] \nonumber \\
& - & \mathbb{E}_{Y Z}[Y u(Y, Z)] -\frac{1}{2} \mathbb{E}_{W}\left[u(W)^{2}\right] \nonumber \\
& =& \left\langle\mathcal{C}_{W T} f-\mathbf{b}, u\right\rangle_{\mathcal{U}}-\frac{1}{2}\left\langle u, \mathcal{C}_{W} u\right\rangle_{\mathcal{U}}. 
\end{eqnarray}
where $\mathbf{b}:=\mathbb{E}_{Y Z}[Y \varphi(Y, Z)] \in \mathcal{U}, \mathcal{C}_{W}:=\mathbb{E}_{W}[\varphi(W) \otimes \varphi(W)] \in \mathcal{U} \otimes \mathcal{U}$ is a covariance operator, and $\mathcal{C}_{W T}:=\mathbb{E}_{W T}[\varphi(W) \otimes \phi(T)] \in \mathcal{U} \otimes \mathcal{F}$ is a cross-covariance operator. 
The generalized least squares solution in RKHS is:
\begin{eqnarray}
f^{*} & = & \arg \min _{f \in \mathcal{F}} \frac{1}{2}\left\langle\mathcal{C}_{W T} f-\mathbf{b}, \mathcal{C}_{W}^{-1}\left(\mathcal{C}_{W T} f-\mathbf{b}\right)\right\rangle_{\mathcal{U}} \nonumber \\
& = & \left(\mathcal{C}_{T W} \mathcal{C}_{W}^{-1} \mathcal{C}_{W T}\right)^{-1} \mathcal{C}_{T W} \mathcal{C}_{W}^{-1} \mathbf{b}
\label{eq:close}
\end{eqnarray}
Eq. (\ref{eq:close}) gives a solution for IV-based regression in closed form. 


\subsubsection{Deep-based Estimator}
\label{sec:deep}

Originally, 2SLS performs linear regressions in both stages under linearity assumption. Recent machine learning methods extend it to non-linear settings with infinite dictionaries of basis functions from reproducing kernel Hibert spaces (RKHS), such as KernelIV \cite{singh2019kerneliv} and DualIV \cite{muandet2019dualiv}. 
Although these methods enjoy desirable theoretical properties, the flexibility of the model is limited, since the basis functions are pre-specified by human-hand or feature engineering\cite{hartford2017deepiv,xu2021learning}.

\quad \\ \noindent \textbf{DeepIV}.\\
DeepIV builds upon deep-based methods, i.e., deep neural networks \cite{hartford2017deepiv}. 
Although there is little theory to justify when learning with neural networks can identify a true model, deep methods make substantially weaker assumptions about the data generating process and automatically learn flexible feature mappings for high-dimension and non-linear data, which saves the human effort selecting pre-defined basis functions and improves the accuracy of causal effect estimation. 
Under additive noise assumption or linearity assumption, \cite{hartford2017deepiv} provide an unique solution for the inverse problem with the learned representation, as follows.

Taking the expectation of both sides of Eq. (\ref{eq:gTY}) conditioned on $\{Z,\mathbf{X}\}$ and applying assumptions formulates the relationship \cite{newey2003instrumental}:
\begin{eqnarray}
\mathbb{E}[Y \mid Z,\mathbf{X}] & = & \mathbb{E}[g(T,\mathbf{X}) \mid Z,\mathbf{X}] + \mathbb{E}[{\epsilon}_Y \mid \mathbf{X}] \nonumber \\
& = &  \int g(T,\mathbf{X}) dF(T \mid Z,\mathbf{X}), 
\end{eqnarray}
where, again, $dF(T \mid Z,\mathbf{X})$ is the conditional treatment distribution obtained from the treatment regression. The relationship defines an inverse problem in structural function identification. Given observational data $\{z_i,\mathbf{x}_i,t_i,y_i\}$, the counterfactual functions are recovered by minimizing the objective:
\begin{eqnarray}
\hat{g} = \text{argmin}_{g \in \mathcal{G}} \sum_{i=1}^n 
\left(y_i - \int_t g(t,x_i) dF(t \mid z_i, \mathbf{x}_i)\right)^2.
\end{eqnarray}

Furthermore, in estimation, DeepIV develops a two-stages procedure. To obtain the the conditional probability estimation $dF(t \mid z_i, \mathbf{x}_i)$ of treatments, deep methods use conditional density estimation model as treatment regression module in stage 1 \cite{darolles2011nonparametric,hartford2017deepiv}. Then, they perform a joint mapping from re-sampled treatments $\hat{T}$ and confounders $\mathbf{X}$ to the counterfactual outcomes $Y$ in stage 2. 

\textbf{Treatment Regression Stage}. Specifically, we use a deep neural network $\pi_\phi(Z,\mathbf{X})$ with parameters $\phi$ to model the conditional density function of treatment $F(T \mid Z,\mathbf{X})$. The objective can be written as:
\begin{eqnarray} \label{eq:deepiv1}
\text{min } \mathcal{L}_1 = l(T, \pi_\phi(Z,\mathbf{X})),
\end{eqnarray}
where $l(T, \pi_\phi(Z,\mathbf{X}))$ would be an $l_2$-loss for continuous outcomes or a log-loss for binary outcomes.
For discrete treatments $T$, we model $\pi_\phi(Z, \mathbf{X})$ with $P(T=k)=\pi_{\phi,k}(Z, \mathbf{X})$ for each treatment arm $T=k$ and where $\pi_{\phi,k}(Z, \mathbf{X})$ is given by the $k$-th element of softmax output in a DNN. For continuous treatments $T$, we model a mixture of Gaussian distributions with component $\pi_{\phi,k}(Z, \mathbf{X})$ and sub-networks $[\mu_{\phi,k}(Z,\mathbf{X})],\sigma_{\phi,k}(Z,\mathbf{X})]$ for Gaussian  distribution parameters $G(\mu,\sigma)$. With enough mixture
components, the network $\pi_\phi(Z,\mathbf{X})$ can approximate arbitrary smooth densities. 

\textbf{Outcome Regression Stage}. We model a counterfactual prediction network $h_\theta$ with parameters $\theta$, to approximate the potential outcome. The objective can be written as:
\begin{eqnarray} \label{eq:deepiv2}
\text{min } \mathcal{L}_2 = \frac{1}{n} \sum_{i=1}^n \left( y_i - \int_t h_\theta(t, x_i) d\hat{F}_{\phi}(t \mid z_i,x_i) \right)^2,
\end{eqnarray}
where $\hat{F}_{\phi}(T \mid Z,\mathbf{X})$ is from the stage 1. Then, we can can optimize the $\hat{F}_{\phi}(T \mid Z,\mathbf{X})$ and $h_\theta(T, \mathbf{X})$ by minimizing the loss $\mathcal{L}_1(\phi)$ and $\mathcal{L}_2(\theta)$ using gradient descent, respectively.

\quad \\ \noindent \textbf{OneSIV}.\\
However, existing deep-based methods require two stages to separately
estimate the conditional treatment distribution and the potential outcome function, which is not sufficiently effective \cite{lin2019one}. Lin et al. \cite{lin2019one} claims that the information from the outcome regression is one significant component for joint distribution of observations, and we should utilizing this information to improve the conditional treatment distribution estimation. 

\textbf{One Stage Regression}. Further, they merge the two stages to leverage the outcome regression $h_\theta(T, \mathbf{X})$ to the
treatment distribution estimation $\hat{F}_{\phi}(T \mid Z,\mathbf{X})$ through a cleverly designed deep neural network structure. Then, they present a joint trade-off objective, as follows: 
\begin{eqnarray}
\text{min}_{\phi,\theta} w_1 \mathcal{L}_1 + w_2 \mathcal{L}_2,
\end{eqnarray}
where $w_1$ and $w_2$ are the hyper-parameters to control the relative importance of treatment regression $\mathcal{L}_1$ (Eq. \ref{eq:deepiv1}) and outcome regression $\mathcal{L}_2$ (Eq. \ref{eq:deepiv2}) obtained from DeepIV. Minimizing this objective, the treatment regression network and the outcome regression network can promote each other's evolution, i.e., Co-evolution.

\quad \\ \noindent \textbf{DFIV}. \\
Combining the theoretical advantages of kernel-based methods and the empirical advantages of deep learning methods, DFIV \cite{xu2021learning} uses deep neural networks (DNNs) to adaptively learn deep features as kernel basis in the 2SLS approach, which fits structural functions with highly nonlinear flexibility. \cite{xu2021learning} develops three DNNs $\{f_{\phi},g_{\xi},u_{\psi}\}$ to learn the corresponding feature mappings for $\{Z,\mathbf{X},T\}$, respectively. Similar to Eqs. (\ref{eq:34})(\ref{eq:35}), we can reformulate the IV-based regression as:
\begin{eqnarray}
u_{\psi,k}(T) = \sum_{i=1}^{d^Z}\sum_{j=1}^{d^X}\alpha_{i,j}^k f_{\phi,i}(Z) g_{\xi,j}(\mathbf{X}) + \epsilon_T,  \\
Y = \sum_{k=1}^{d^T}\sum_{j=1}^{d^X}\beta_{k,j} u_{\psi,k}(T) g_{\xi,j}(\mathbf{X}) + \epsilon_Y, 
\end{eqnarray}
where $f_{\phi,i}(Z)$ denotes the $i$-th element in the outcome vector of instrument representation network $f_{\phi}(Z)$, $g_{\xi,j}(\mathbf{X})$ is the $j$-th element in the outcome vector of covariate representation network $g_{\xi}(\mathbf{X})$, and $u_{\psi,k}(T)$ is the $k$-th element in the outcome vector of treatment representation network $u_{\psi}(T)$. $\{d^Z,d^X,d^T\}$ denotes the dimension of the outcome vector $f_{\phi}(Z)$, $g_{\xi}(\mathbf{X})$, and $u_{\psi}(T)$. $\mathbf{A}= [\alpha_{i,j}^k]_{i,j,k}$ and $\mathbf{B}=[\beta_{i,j}]_{i,j}$ denote the corresponding coefficients in the linear associations between features $\{f_{\phi}(Z),g_{\xi}(\mathbf{X}),u_{\psi}(T), Y\}$. 

\textbf{Treatment Regression Stage}.  Fixing the parameter $\psi$ of the treatment representation network $u_{\psi}(\cdot)$ and the parameter $\xi$ of the covariate representation network $g_{\xi}(\cdot)$ during stage 1, DFIV aims to regress the conditional expectation $\mathbb{E}[u_{\psi}(T) \mid f_{\phi}(Z) \otimes g_{\xi}(\mathbf{X})]$ by learning the network parameter $\phi$ and the coefficient matrix $\mathbf{A} \in \mathbb{R}^{d^T \times (d^Z \cdot d^X)}$, where $f_{\phi}(Z) \otimes g_{\xi}(\mathbf{X})$ denotes the multiplication combination set $[f_{\phi,i}(Z) g_{\xi,j}(\mathbf{X})]_{i,j}$. 
\begin{eqnarray}
\phi^*
& = &\text{argmin}_{\phi} \mathcal{L}_1(\phi), \\
\Scale[0.8]{\mathcal{L}_1(\phi) } & =  & 
\Scale[0.8]{ \frac{1}{n} 
\sum_{i=1}^{n} \left[ 
\left\|u_{\psi}(t_i)-\mathbf{A} f_{\phi}(z_i) \otimes g_{\xi}(x_i) \right\|^{2}
+ \lambda_{1}\|\mathbf{A}\|^{2}  \right]}
\\
\mathbf{A}(\phi) & = & u_{\psi}(T)^{\prime} \mathbf{C} (\mathbf{C}^{\prime} \mathbf{C} + n \lambda_1 I)^{-1}
\end{eqnarray}
To simplify notation, in this section, we denotes by $\mathbf{C}=f_{\phi}(Z) \otimes g_{\xi}(\mathbf{X}) \in \mathbb{R}^{n \times (d^Z \cdot d^X)}$. We can then learn the parameters $\phi$ of the instrument representation network $f_{\phi}(\cdot)$ by minimizing the loss $\mathcal{L}_1(\phi)$ using gradient descent.

\textbf{Outcome Regression Stage}.  Fixing the parameters $\phi$ of the instrument representation network $f_{\phi}(\cdot)$ and the parameter $\xi$ of the covariate representation network $g_{\xi}(\cdot)$ during stage 2, DFIV predicts the structural function $\mathbb{E}[Y \mid u_{\psi}(T) \otimes g_{\xi}(\mathbf{X})]$ by learning the network parameter $\psi$ and the coefficient matrix $\mathbf{B} \in \mathbb{R}^{1 \times (d^T \cdot d^X)}$. To simplify notation, in this section, we use $\mathbf{D}=f_{\psi}(T) \otimes g_{\xi}(\mathbf{X})$ denotes the multiplication combination set $[f_{\psi, k}(T) g_{\xi,j}(\mathbf{X})]_{k,j}$. 
\begin{eqnarray}
\psi^*
& = &\text{argmin}_{\psi} \mathcal{L}_2(\psi), \\
\Scale[0.8]{\mathcal{L}_2(\psi) } & =  & 
\Scale[0.8]{ \frac{1}{n} 
\sum_{i=1}^{n} \left[ 
\left\|y_i-\mathbf{B} f_{\psi}(t_i) \otimes g_{\xi}(x_i) \right\|^{2}
+ \lambda_{2}\|\mathbf{B}\|^{2}  \right]}
\\
\mathbf{B}(\psi) & = & Y^{\prime} \mathbf{D} (\mathbf{D}^{\prime} \mathbf{D} + n \lambda_2 I)^{-1}
\end{eqnarray}
We can then learn the parameters $\psi$ of the treatment representation network $f_{\psi}(\cdot)$ by minimizing the loss $\mathcal{L}_2(\psi)$ using gradient descent.

Note that the covariate representation network $g_{\xi}(\cdot)$ is fixed during stage 1 and stage 2. To update the covariate network $g_{\xi}(\cdot)$, fixing the parameters $\psi$ and $\phi$, we minimize the loss $\mathcal{L}_1(\xi) + \mathcal{L}_2(\xi)$  using gradient descent. Then, we adopt an alternating training strategy to iteratively optimize the representations for $g_{\psi}(\cdot)$, $g_{\phi}(\cdot)$ and $g_{\xi}(\cdot)$.

\subsubsection{GMM-based Estimator}
\label{sec:gmm}

In the presence of heteroskedasticity, although the counterfactual function estimation of the standard IV estimators and some variants is consistent with the true potential outcomes, the standard errors are inconsistent, preventing valid inference \cite{baum2003instrumental}. 
Assuming observational data can be formalized in moment conditions, when facing heteroskedasticity of unknown form, we can make use of the conditional moment restrictions to allow for efficient estimation. That is, instrumental variable regression and 2SLS can be seen as special cases of generalized method of moments (GMM), introduced by \cite{hansen1982large}, which is a prototypical (non-)parametric estimator \cite{hansen2000testing, wooldridge2010econometric, hayashi2011econometrics}. 



The standard IV estimator is a special case of GMM. Satisfying the IV assumptions, the instruments $Z$ is correlated with the endogenous treatments $T$ and orthogonal to the unmeasured confounders ${\epsilon}_T$/$\epsilon_Y$ at the same time, i.e., $Z \perp U$. Then, we can design a IV-based GMM estimator to satisfy the orthogonality conditions with the overidentified context. Under the additive noise assumption (Eq. (\ref{eq:fZX})(\ref{eq:gTY})), the moment conditions for instruments $Z \in \mathbb{R}^{n \times d^Z}$ can be formulated as $\mathbb{E}[Z\epsilon_T]=\mathbb{E}[Z\epsilon_Y]=0$. The $d^Z$ instruments give a set of $d^Z$ moments:
\begin{eqnarray}
\hspace{-18pt} l_i(g) & = & z_i^{\prime} u_i = z_i^{\prime} (y_i - g(t_i,x_i)), i=1,\cdots,n \\
\hspace{-18pt} \mathbb{E}[l(g)] & = & \frac{1}{n} \sum_{i=1}^nl_i(g) = \frac{1}{n} \sum_{i=1}^nz_i^{\prime} (y_i - g(t_i,x_i)) = \mathbf{0}.
\end{eqnarray}
where $\mathbb{E}[l(g)]$ is a $d^Z$ vector, and we set $L_j = \mathbb{E}[l(g)]_j$ to denote the $j$-th element in the expectation error vector $\mathbb{E}[l(g)]$. The intuition of GMM is to choose an estimator for function $g$, and set these $d^Z$ moments as close to zero as possible. 

In the estimation of potential outcome function, if the number of unknown parameter is exactly $d^Z$, the estimated equation is exactly identified —— the $d^Z$ moment conditions and the $d^Z$ parameters in regression function. If we have less unknown parameters than conditional moment restrictions, then the estimated equation is overidentified, and we cannot find a prediction function $g$ to set all $d^Z$ sample moment conditions $[L_j = \mathbb{E}[l(g)]_j]_{j = 1,\cdots,d^Z}$ to exactly zero. Thus, GMM estimator replace the theoretical expected value $\mathbb{E}[\cdot]$ with its empirical analog—sample average:
\begin{eqnarray}
\hspace{-18pt} \mathcal{J}(\theta) = \sum_{j=1}^{d^Z} L_j^2(\theta) = \| L(\theta) \|^2 = L(\theta)^{\prime} W L(\theta) = \sum_{j=1}^{d^Z} [l(g_\theta)]_j]^2.
\label{eq:72}
\end{eqnarray}
where $W=I$ is an identify matrix, meaning the average effect. 
Then we minimize the norm of this expression with respect to function $g_\theta$. The minimizing function of $g_\theta$ is our estimate for $g$.

Although GMM is an incredibly flexible estimator, in practical, there are an infinite number of moment conditions with IV independence assumptions. Imposing all of them is infeasible with finite data. Therefore, recent literature proposes a series of minimax approaches to reformulate the minimax optimization problem.

\quad \\ \noindent \textbf{Minimax Approachs}. \\
There has also been a recent surge in interest with minimax approaches that reformulate conditional moment conditions as a minimax optimization problem. For example, Lewis \& Syrgkanis (2018); Zhang et al. (2020) use the reformulation $\sup _{h \in L_{2}\left(Z_{i}\right)}\left(\mathbb{E}\left[h\left(Z_{i}\right)\left(Y_{i}-f^{*}\left(T_i, \mathbf{X}_{i}\right)\right)\right]\right)^{2}$. Bennett et al. (2019), Bennett \& Kallus (2020), Muandet et al. (2020), while Dikkala et al. (2020), Chernozhukov et al. (2020), and Liao et al. (2020), employ other reformulations, i.e., $\sup _{h \in L_{2}\left(Z_{i},\mathbf{X}_{i}\right)}\left(\mathbb{E}\left[h\left(Z_{i},\mathbf{X}_{i}\right)\left(Y_{i}-f^{*}\left(T_i, \mathbf{X}_{i}\right)\right)\right]\right)^{2}$.

With the rapid development of machine learning algorithms, researchers apply adaptive non-parametric learners such as reproducing kernel Hilbert spaces, random forests, and neural networks to reformulate GMM estimation to the minimax optimization problem \cite{lewis2020agmm, bennett2019deep, zhang2020maximum}. 
In machine learning and statistics, researchers formulate the target estimand as an objective minimization problem. Then, Lewis et al. \cite{lewis2020agmm} formulate the expectation minimization problem as the maximum moment deviation over the set of potential functions, refered as Adversarial GMM (AGMM):
\begin{eqnarray}
h^* = \text{arginf}_{h \in \mathcal{H}} \text{sup}_{f \in \mathcal{F}} \mathbb{E}[(Y - h(T,\mathbf{X}))f(Z,\mathbf{X})].
\end{eqnarray}
Similar to Wasserstein and MMD GANs \cite{arjovsky2017wasserstein,li2017mmd}, the formulation proposes a learner network $h$ to set moments as close to zero as possible, and an adversary network $f$ to identify moments that are violated for the chosen $h$.
\cite{lewis2020agmm} offers main theorems and applications for several hypothesis spaces of practical interest including reproducing kernel Hilbert spaces (RKHS), functions defined via shape restrictions, random forests, and neural networks. 

Given observational data $\{z_i,x_i,t_i,y_i\}_{i=1,\cdots,n}$, to obtain optimal $h_{\phi}$ and $f_{\psi}$, AGMM \cite{lewis2020agmm} minimizes the empirical analogue of the minimax objective:
\begin{eqnarray}
\phi^* & = & \text{arginf}_{\phi \in \Phi} \text{sup}_{\psi \in \Psi} \mathbb{E}[(Y - h_{\phi}(T,\mathbf{X}))f_{\psi}(Z,\mathbf{X})] \nonumber \\
& - & \lambda_1 \| \psi \|^2 - \mathbb{E}[f_{\psi}(Z,\mathbf{X})^2] + \lambda_2 \| \phi \|^2.
\end{eqnarray}
where $\{\lambda_1,\lambda_2\}$ are the hyper-parameters for penalty items $ \| \phi \|^2$ and $ \| \psi \|^2$. 

\quad \\ \noindent \textbf{DeepGMM}. \\
With infinite moment conditions, using identify matrix $I$ as unweighted vector norm can lead to significant inefficiencies in the minimization of objective Eq. (\ref{eq:72}) \cite{hansen1982large,hansen1996finite}. \cite{hansen1982large,hansen1996finite} claim that weighting moment conditions by their inverse covariance would yield minimal variance estimates, and it is sufficient to consistently estimate this covariance.
Based on the optimally weighted Generalized Method of Moments (GMM) \cite{hansen1982large,bennett2019deep,bennett2020variational}, DeepGMM \cite{bennett2019deep} construct an optimal combination of moment conditions via adversarial training, with the objective:
\begin{eqnarray}
\phi^* & = & \text{arginf}_{\phi \in \Phi} \text{sup}_{\psi \in \Psi} \mathbb{E}[(Y - h_{\phi}(T,\mathbf{X}))f_{\psi}(Z,\mathbf{X})] \nonumber \\
& - & \frac{1}{4} \mathbb{E}[(Y - h_{\phi}(T,\mathbf{X}))^2 f_{\psi}^2(Z,\mathbf{X})].
\end{eqnarray}
Notably, DeepGMM \cite{bennett2019deep} has a few tuning parameters: the models $\mathcal{F}$ and $\mathcal{H}$ (i.e., the neural network architectures) and whatever parameters the optimization method uses.
Besides, other reformulations of minimax problem are developed by \cite{liao2020provably,chernozhukov2020adversarial}

\subsubsection{Confounder Balance Estimator}

\begin{figure}
\begin{center}
\includegraphics[width=0.72\linewidth]{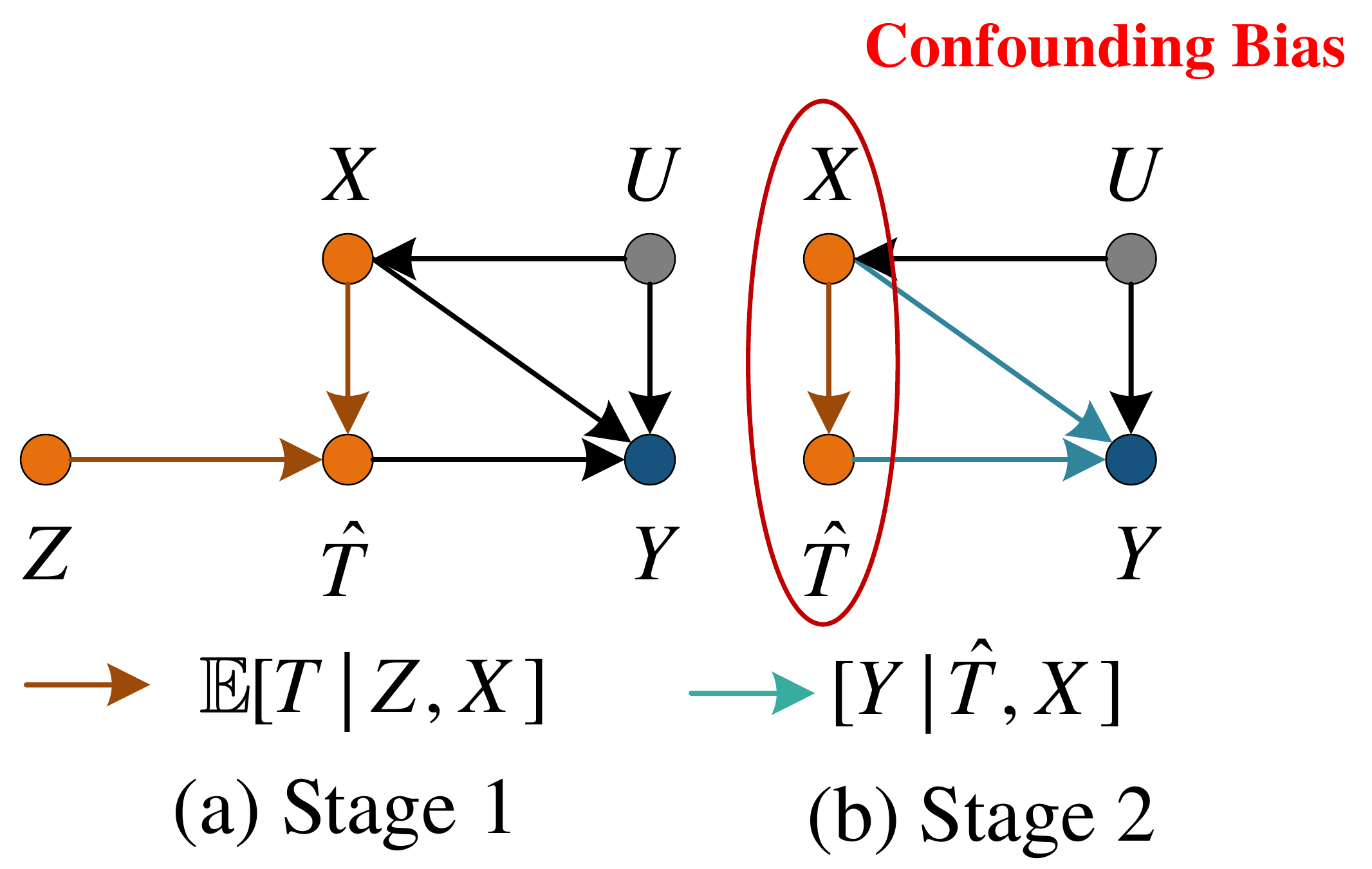}
\caption{Confounding bias from observed confounders.}\label{fig:confounding}
\end{center} 
\end{figure}

With the development of machine learning, instrumental variables are no longer limited to simple linear models. The recent IV models described above have focused on various complex setting, where interactions between various variables may exist, such as $T=ZX + X + U$.
At this point, if we do not consider the joint effect of modeling covariates and IVs, the effect of IVs on the treatment variables will be very limited, i.e., weak IV. Therefore, these algorithms combine observed confounders and IVs to predict the conditional distribution of the treatments to eliminate unmeasured confounding bias in stage 1. However, this introduces additional bias due to imbalanced covariates $X$ on different treatment arms in stage 2 (Fig.~\ref{fig:confounding}). 

\quad \\ \noindent \textbf{CBIV}. \\
Wu et al. \cite{wu2022instrumental} focus on treatment effect estimation with IV regression under homogeneity assumptions, and they propose a Confounder Balanced IV Regression (CB-IV) algorithm to further remove the confounding bias from observed confounders by balancing in nonlinear scenarios.

Based on the Homogeneous Instrument-Treatment Assumption, Wu et al. \cite{wu2022instrumental} model a more general causal relationship by relaxing the additive assumption to multiplicative assumption on response-outcome function as:
\begin{eqnarray}
\label{complicated_function1}
    & \Scale[0.95]{T}=\Scale[0.95]{f_1(Z,X)+f_2(X,U)} \\ 
    & \Scale[0.95]{Y}=\Scale[0.95]{g_1(T,X)+g_2(T)g_3(U)+g_4(X,U), Z \perp U,X}
\label{complicated_function2}
\end{eqnarray}
where ${f_{i}}(\cdot),{g_{j}}(\cdot)$ are unknown and potentially non-linear continuous functions. $g_2(T)g_3(U)$ denotes the multiplicative terms of $U$ with $T$ (e.g., $U^2T-UT+U$). 
The completeness of $\mathbb{P}({T \mid Z,X})$ and $\mathbb{P}({Y \mid T,X})$ guarantees uniqueness of the solution \cite{newey2003instrumental}.

The CB-IV algorithm contains the following three main components: 

\textbf{Treatment Regression in Stage 1:} For continuous treatment $T$, CBIV regresses treatment $T$ with IVs $Z$ and observed confounders $X$. 
\begin{eqnarray}
\label{eq:loss_Tcont2}
\mathcal{L}_T = \frac{1}{n} \sum_{i=1}^{n}\sum_{j=1}^{m} \left(t_i - \hat{t}_i^j\right)^2, \hat{t}_i^j \sim \hat{P}(t_i|z_i, x_i), 
\end{eqnarray} 
we sample $m$ (the larger the better) treatment $\{\hat{t}_i^j\}_{j=1,...,m} $ for each unit $\{z_i,x_i\}$ to approximate the true treatment $t_i$. 
Empirically, the above objective (Eq. (\ref{eq:loss_Tcont2})) is sufficient to accurately estimate causal effects in continuous CB-IV framework. 

\textbf{Confounder Balance in Stage 2:} 
For continuous treatment $T$, we learn a ''balanced'' representation (i.e., $C$) of the observed confounders $X$ as $C=f_\theta(X)$ via mutual information (MI) minimization constraints: firstly, we use variational distribution $Q_{\psi}(\hat{T} \mid C) = \mathcal{N}(\mu_{\psi}(C),\sigma_{\psi}(C))$ parameterized by neural networks $\{\mu_{\psi}, \sigma_{\psi}\}$ to approximate the true conditional distribution $P (\hat{T} \mid C)$; then, we minimize the log-likelihood loss function of variational approximation $Q_{\psi}(\hat{T} \mid C)$ with $n$ samples to estimate MI:
\begin{eqnarray}
{\text{ disc}(\hat{T}, C) =\frac{1}{n^2}\sum_{i=1}^{n}\sum_{j=1}^{n}\left[\log Q_{\psi}\left(\hat{t}_{i}\mid c_{i}\right)-\log Q_{\psi}\left(\hat{t}_{j}\mid c_{i}\right)\right]}. 
\end{eqnarray}
where, $C=f_\theta(X)$. We adopt an alternating training strategy to iteratively optimize $Q_{\psi}(\hat{T} \mid C)$ and the network $C=f_{\theta(X)}$ to implement balanced representation in the Confounder Balancing. 

\textbf{Outcome Regression:} Finally, we propose to regress the outcome with the estimated treatment $\hat{T} \sim P(T|Z, X)$ obtained in treatment regression module and the representation of confounders $C=f_\theta(X)$ obtained in confounder balancing module:  
\begin{eqnarray}
\mathcal{L}_Y=\Scale[1.0]{\frac{1}{n} \sum\limits_{i=1}^n \left
(y_i - h_\xi(\hat{t}_i,f_\theta(x_i))\right)^2}
\label{tenfourcont}
\end{eqnarray}
where $\hat{t}_i \sim \hat{P}(T|Z, X)$ and $f_\theta(x_i)$ are derived from treatment regression module and confounder balancing module, respectively.

Theoretically and empirically, CBIV confirms that eliminating confounding bias in the outcome regression stage will contribute to more accurate treatment effect estimation. 

\subsection{Limitation and Future Work}

\subsubsection{Limitation}

\textbf{Invalid IV.} The above methods are reliable only if the pre-defined IVs are valid and strongly correlated with the treatment variable. However, such valid IVs are hardly satisfied due to the untestable exclusion association with outcome \cite{wu2022instrumental}. Therefore, we have to rely on expert knowledge to select the instrumental variables, but this often does not guarantee the validity of the instrumental variables: IV does not have a direct effect on the outcome variable, only indirectly through the treatment variable. As an alternative, in instrumental variable literature, researchers usually implement Randomized Controlled Trials (RCTs) to obtain exogenous IVs, such as Oregon health insurance experiment \cite{finkelstein2012oregon} and effects of military service on lifetime earnings \cite{angrist1990lifetime}, which are too expensive to be universally available. 

\noindent \textbf{Weak IV and Mis-specified Model.} In the real world, ones always consider a large number of variables (i.e., pre-treatment variables) that are relevant to the outcome and then choose treatments in the hope of obtaining the optimal results. 
The instrumental variables are usually only a few, or even non-existent. Besides, the potential mechanisms of data generation are complex, and there may be interactions between various variables, such as $T=ZX + X + U$. In other words, IVs may have little causal effect on the treatment variables, which we call weak IV. Therefore, machine learning algorithms tend to combine observed confounders and IVs to predict the conditional distribution of the treatments to eliminate unmeasured confounding bias. Wu et al. \cite{wu2022instrumental} points out that these methods would make the predicted 
treatments $\hat{T}$ correlate with the observed variables $X$ and imbalanced variables $X$ will bring additional confounding bias for outcome regression, if the outcome model is misspecified (Fig.~\ref{fig:confounding}).

\noindent \textbf{Limited Sample.} Machine learning algorithms are data-driven algorithms, and their performance is highly dependent on the number of samples. When the sample size is infinite, we can obtain unbiased estimates by the above algorithm. However, in finite samples, machine learning algorithms are prone to overfitting, leading to errors in the regression of the intervening variables, which will further lead to failure in the regression of the resulting coefficients. In addition, imbalanced covariates can also induce overfitting and introduce sample selection bias.

\subsubsection{Future Work}

\noindent \textbf{Causal Discovery.} When we have access to a large number of variables, we can try to mine the instrumental variables from the data by using causal discovery algorithms with latent variable, including constraint-based methods, score-based methods and model-based methods, such as SCORE \cite{Score2022}.

\noindent \textbf{Generalized Method of Moments.} 
GMM is an incredibly flexible IV estimator that relies on a large number of moment conditions with IV independence conditions. 
With the advancement of machine learning algorithms, nonlinear independence detection algorithms have also been developed, which has outperformed first-order moment independent etc.
Therefore, a natural idea is to use independent testing algorithms instead of moment conditions to constrain the instrumental variable regression, such as HSIC-X \cite{HSICX2022}

\noindent \textbf{Confounder Balance.}
In the presence of unmeasured confounders and the above IV methods raises a very interesting bias problem in non-linear IV methods. These methods would suffer from the bias from the observed confounders, which are imbalanced in the second stage of IV regression. To address this problem, CBIV \cite{wu2022instrumental} proposes a confounder balanced IV regression algorithm by a novel combination of the confounder balancing and IV regression, where the confounder balancing is designed for removing the bias from the observed confounders and the IV regression is for removing the bias from the unobserved variables. In the provided theoretical analyses and numerical experiments, \cite{wu2022instrumental} demonstrates the effectiveness of the proposed algorithm. In the future, confounder balance is an issue that has to be considered in instrumental variable regression.

\section{Control Function}
\label{sec:cfn}

Another statistical method to correct for unmeasured confounding bias is control function (CFN), also know as two-stage residual inclusion. The principle of control function can be traced back to some early works\footnote{Based on \cite{wooldridge2015control}.} \cite{telser1964iterative,goldberger2008selection}, a control function is a variable that renders known cause variables (i.e., Treatments) appropriately exogenous in the outcome regression \cite{barnow1981selection,goldberger2008selection,cameron2005microeconometrics}. In observational data, conditional on control function or confounders\footnote{Under the unconfoundedness assumption, the role of control function in regression is consistent with that of confounding variables}, CFN estimator makes the treatment appropriately exogenous in the regression queation. CFN is a two-stage residual inclusion method, which deponds on the parameters estimated by treatments $T$ and valid IVs $Z$ in stage 1 \cite{heckman1985alternative}. And, it is not only useful in linear cases, but also in the non-linear scenarios to elimate bias for endogeneity.  

\begin{figure}
\begin{center}
\includegraphics[width=0.72\linewidth]{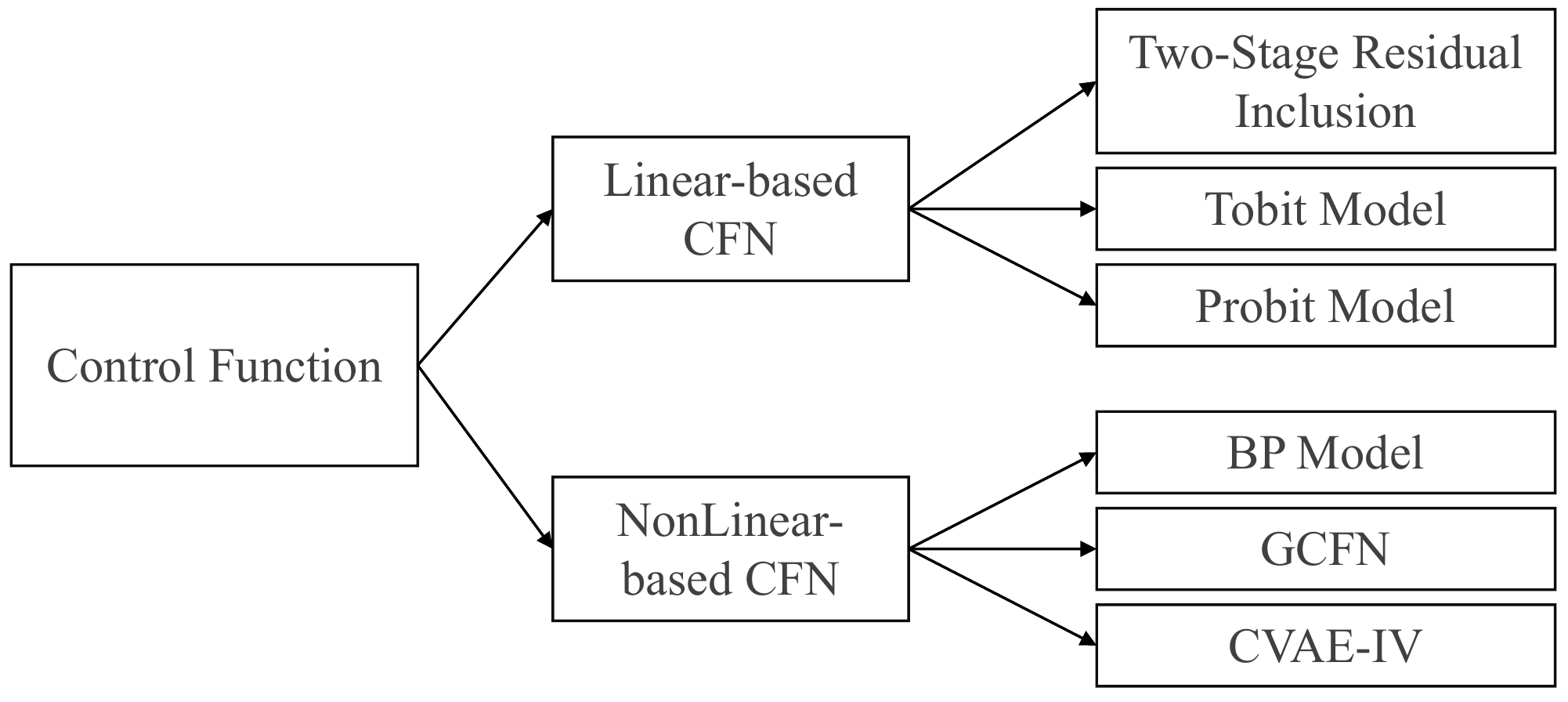}
\caption{Categorization of Control Function Estimators.}\label{fig:mlcfn}
\end{center} 
\end{figure}

In stage 1, based on the variation induced by exogenous IVs in the treatment regression from IVs $Z$ to treatments $T$, we can obtain a generalized residual that serves as control function. As for stage 2, conditional on control function estimated in stage 1, the treatment become appropriately exogenous in the outcome regression. Next, we show how CFN regression works in cansal inference and machine learning, including linear and non-linear scenarios, as shown in Fig. \ref{fig:mlcfn}. 

\subsection{Linear-based CFN}
\label{sec:linearCFN}

\subsubsection{Control Function Estimations}
For the most part, the usage of CFN maintains the spirit of the earlier definitions and estimations \cite{wooldridge2015control}. In the presence of unmeasured confounders $\mathbf{U}$, we assume $\mathbf{V}=f({\mathbf{U}})$ as unmeasured noise for treatments and model structural linearity function in constant coefficients:
\begin{eqnarray}
\label{eq:linearCC_ZT}
{T} & = & {Z} \alpha + f({\mathbf{U}}) = {Z} \alpha + \mathbf{V}, \\
\label{eq:linearCC_TY}
{Y} & = & {T} \beta + {\mathbf{U}} = {T} \beta + f^{-1}(\mathbf{V}), 
\end{eqnarray}
where instrumental variables are independent of unmearsured confoudners, i.e., $\mathbb{E}(Z\mathbf{U})=0$ and $\mathbb{E}(\mathbf{U} \mid Z)=\mathbb{E}(\mathbf{U})$. Similarity, the IVs are uncorrelated with $f(\mathbf{U})$. In linearity, we model the $\mathbf{U}$-$T$ association (i.e., the residuals) as $\mathbf{V} = f({\mathbf{U}}) = \mathbf{U} / \rho$, and $f^{-1}$ is the inverse function of association $f$. Then we can obtain: 
\begin{eqnarray}
\label{eq:linearCC_UT}
f^{-1}(\mathbf{V}) = \rho \mathbf{V},
\end{eqnarray}
where $\rho$ is the population regression coefficient. We plug it into the Eq. (\ref{eq:linearCC_TY}):
\begin{eqnarray}
\label{eq:linearCC_ETY}
{Y} = {T} \beta + \rho \mathbf{V}.
\end{eqnarray}

In the observational data $\mathcal{D} = \{Z,\mathbf{U},T,Y\}$, we do not observe $\mathbf{U}$ or the residuals $\mathbf{V} = f({\mathbf{U}})$. Nevertheless, based on Eq. (\ref{eq:linearCC_ZT}), we can get $\mathbf{V} = T - {Z} \alpha$. Because $Z$ is uncorrelated with $\mathbf{V}$ in the linear model, we can consistently estimate the coefficient $\alpha$ by OLS. The two-step control function procedure is as follows:

\noindent \textbf{The Residual Learning Stage:} in stage 1 of CFN, we perfrom the regression of the treatments $T$ on exogenous IVs $Z$:
\begin{eqnarray}
& \hat{\alpha} = \left({Z}^{\prime} {Z}\right)^{-1} {Z}^{\prime} {T} 
= \left({Z}^{\prime} {Z}\right)^{-1} {Z}^{\prime} ({Z} \alpha + \mathbf{\mathbf{V}}) = \alpha, \\
& \hat{\mathbf{\mathbf{V}}} = {T} - {Z} \hat{\alpha} = \mathbf{V}.
\end{eqnarray}

\noindent \textbf{The Outcome Regression Stage:} in stage 2 of CFN, based on the association between residuals $\mathbf{V}$ and unmeasured confounders $\mathbf{U}$, we can regard residuals $\mathbf{V}$ as a control function for unmeasured confounders. Then we can control the residuals $\mathbf{V}$ to estimate the conditional average causal effect of treatments $T$ on outcomes $Y$:
\begin{eqnarray}
\text{CATE} = \mathbb{E}[Y(T=t) - Y(T=0) \mid \mathbf{V}]. 
\end{eqnarray}
or dose-response function (ITE):
\begin{eqnarray}
\text{ITE} = Y(T=t,\mathbf{V}) - Y(T=0,\mathbf{V}). 
\end{eqnarray}

The coefficents on $Z$ and $T$ from CFN estimator arenumerically identical to that of 2SLS estimator \cite{hausman1978specification}. In above linear setting, CFN estimator does not lead to a novel estimator different from 2SLS. In fact, if we perform OLS in the outcome regression, we find it is hard to obtain unbiased causal effect and we need to control the CFN/confounders.

\subsubsection{Binary/Discrete treatment effects}
\textbf{Binary/Discrete Treatment} $T = \{0,1\}$ is a special case for CFN. When the treatment is a binary random variable, that is also applicable to discrete variables, a choice is to utilize the binary nature of treatment $T$ and replace the linear regression with a binary response model. The structural equation is supplemented with the continuous models in Eqs. (\ref{eq:linearCC_ZT}) and (\ref{eq:linearCC_TY}):
\begin{eqnarray}
\label{eq:binaryCC_ZT}
{T} & = & \mathbbm{1} \{ {Z} \alpha + \mathbf{V} > 0 \}, \\
\label{eq:binaryCC_TY}
{Y} & = & {T} \beta + \mathbf{U}, 
\end{eqnarray}
where $\mathbbm{1}\{\cdot\}$ is the indicator function, and $\{\mathbf{U}, \mathbf{V} \}$ are independent of $Z$. There is a linear causal relationship between $\mathbf{U}$ and $\mathbf{V}$. Without loss of generality, we assume that the residual satisfies $\mathbf{V} \sim \mathcal{N}(0,1)$. Thus, the treatment assignment can be regarded as a probit model: 
\begin{eqnarray}
\label{eq:probit}
P(T=1 \mid Z) = \Phi(Z \alpha),
\end{eqnarray}
where $\Phi(\cdot)$ is the standard normal cumulative distribution function. Then we can derive a CFN for binary treatment cases \cite{petrin2010control,wooldridge2010econometric}. 

In stage 1, we estimate the probit model in Eq. (\ref{eq:probit}) and obtain the \emph{generalized residual}:
\begin{eqnarray}
\hat{r_\mathbf{V}} = T \lambda (Z \alpha) - (1-T) \lambda (- Z \alpha)
\end{eqnarray}
where $\lambda(\cdot) = \frac{\phi}{\Phi}(\cdot)$ is the well-known inverse Mills ratio \cite{greene2003econometric}.

In stage 2, we control the generalized residual $r_\mathbf{V}$ to estimate the conditional average causal effect of treatments $T$ on outcomes $Y$:
\begin{eqnarray}
\label{eq:binaryCATE}
\text{CATE} = \mathbb{E}[Y(T=t) - Y(T=0) \mid r_\mathbf{V}]. 
\end{eqnarray}
One limitation for CFN in binary/discrete treatment cases is that the results is reliable only when the designed probit model for $T$ is correct. If the probit model is correctly specified, then the CFN estimator would give an unbias causal effect. 

\subsubsection{Heterogeneous treatment effects}
When the coefficients in the structural function is correlated the treatment variable, there are heterogeneous treatment effects in observational data. The random coeffient setting is called a "correlated random coefficient" (CRC) model \cite{heckman1998instrumental,card2001estimating}. Consider the outcome structural function as:
\begin{eqnarray}
\label{eq:heterogeneous_T}
T = Z \alpha + \mathbf{V}, \\
\label{eq:heterogeneous_Y}
Y = T U_1 + U_2, 
\end{eqnarray}
where all unobservables are independent of IVs, i.e., $Z \perp \{U_1, U_2, \mathbf{V}\}$, and the unobservables $U_1$ and $U_2$ are linearly correlated with the residual $\mathbf{V}$:
\begin{eqnarray}
\mathbb{E}[U_1 \mid \mathbf{V}] = \eta \mathbf{V} + \beta, \mathbb{E}[U_2 \mid \mathbf{V}] = \psi \mathbf{V} + c, 
\end{eqnarray}
where $\{ \beta, c \}$ are constant terms, and $\{ \eta, \psi \}$ are the corresponding regression coefficients. 

In the heterogeneous treatment effects dataset, there are two sources of unmeasured confounding bias from $U_1$ and $U_2$. In this cases, we focus on the average treatment effect, i.e., $\beta = \mathbb{E}(U_1)$. Then, we set $U_1 = \mathbb{E}(U_1) + R, \mathbb{E}(R)=0$, and reformulate the outcome structural function as:
\begin{eqnarray}
\label{eq:heterogeneous_Y2}
Y & = & T \beta + T R + U_2, 
\end{eqnarray}
where $\mathbb{E}[R] = \eta \mathbf{V}$ and the correlation between $T$ and $R$ satisfies the assumption: $\text{Cov}(T,R \mid Z) = \text{Cov}(T,R)$ \cite{card2001estimating}. Then we formulate the CFN estimator as:
\begin{eqnarray}
\mathbb{E}[Y(T) \mid U_1, U_2] & = & \mathbb{E}[Y(T) \mid \mathbf{V}, T \mathbf{V}] \nonumber \\
& = & T \beta + \eta T \mathbf{V} + \psi \mathbf{V} + c. 
\end{eqnarray}

In stage 1, we regress the treatment $T$ on the exogenous IVs $Z$:
\begin{eqnarray}
\hat{\alpha} = \left({Z}^{\prime} {Z}\right)^{-1} {Z}^{\prime} {T} 
= \left({Z}^{\prime} {Z}\right)^{-1} {Z}^{\prime} ({Z} \alpha + \mathbf{\mathbf{V}}) = \alpha
\end{eqnarray}
Thus, the residual is.
\begin{eqnarray}
\hat{\mathbf{\mathbf{V}}} = {T} - {Z} \hat{\alpha} = \mathbf{V}.
\end{eqnarray}

In stage 2, we control the residual $\mathbf{V}$ and the multiplicative interaction $T \mathbf{V}$ to estimate the conditional average causal effect of treatments $T$ on outcomes $Y$:
\begin{eqnarray}
\text{CATE} = \mathbb{E}[Y(T=t) - Y(T=0) \mid \mathbf{V}, T \mathbf{V}]. 
\end{eqnarray}
Similar CFNs are also applicable to discrete treatment cases. 

\subsection{NonLinear-based CFN}
\label{sec:nonlinearCFN}

In the previous section, we have introduced contron function methods employed for linear models, including Probit and Tobit. \cite{smith1986exogeneity,rivers1988limited, wooldridge2015control} broaden the scope of the CFN applications. Here, we detail the flexibility of the CFN estimator in the complex non-linear models using machine learning methods.

Consider a simple nonlinear model (observed confounders $\mathbf{X}$ includes a multiplicative interaction $T\mathbf{X}$):
\begin{eqnarray}
\label{eq:nonlinearCC_ZT}
T & = & Z \alpha_1 + \mathbf{X} \alpha_2 + \mathbf{V}, \\
\label{eq:nonlinearCC_TY}
Y & = & \mathbf{X} \beta_1 + T \mathbf{X} \beta_2 + \mathbf{U}, \mathbf{U} = \mathbf{V} \rho. 
\end{eqnarray}
According to the IV's three conditions, we have that $Z \perp \{\mathbf{X},\mathbf{U},\mathbf{V}\}$. In this model, the treatment is continuous, then we obtain the residual in the stage 1. 
\begin{eqnarray}
\hat{\mathbf{V}} = T - \mathbb{E}[T \mid Z, \mathbf{X}] = T - (Z, \mathbf{X})(\hat{\alpha_1}, \hat{\alpha_2})^{\prime} = \mathbf{V}
\end{eqnarray}
where $(Z,\mathbf{X})$ denotes the joint vector of $Z$ and $\mathbf{X}$, and $(\hat{\alpha_1}, \hat{\alpha_2})$ is the corresponding coefficients. Sequentially, we can perform the outcome regression on $\mathbf{V}$, $\mathbf{X}$, and the interaction $T \mathbf{X}$:
\begin{eqnarray}
\text{CATE} = \mathbb{E}[Y(T=t) - Y(T=0) \mid \mathbf{V}, \mathbf{X}, T\mathbf{X}]. 
\end{eqnarray}
A similar estimator can be built for a discrete treatment case, in the discrete model Eq. (\ref{eq:binaryCATE}) \cite{terza2008two,petrin2010control}. The limitation is that the results are reliable only when we have modeled the correct model for non-linear relationship with the prior knowledge of interaction $T \mathbf{X}$. In the next section, we will give a general solution through probit models.

\subsubsection{Non-Parametric BP Estimator}
For more general models, there may be some more complex non-linear relationship in the causal structural function. Based on the probit model \cite{rivers1988limited}, Blundell and Powell (BP) \cite{blundell2003endogeneity} proposes a non-parametric extension of the Rivers-Vuong approach \cite{rivers1988limited}, which is applicable in most general setting:
\begin{eqnarray}
\label{eq:BP_ZT}
T & = & f(Z,\mathbf{X}) + \mathbf{V}, \\
\label{eq:BP_TY}
Y & = & g(\mathbf{X},T,\mathbf{U}). 
\end{eqnarray}
where $f(\cdot)$ and $g(\cdot)$ are the structural functions. The target of BP approach is to estimate the Average Structural Function (ASF) of outcome, defined as follows:
\begin{eqnarray}
\text{ASF}(\mathbf{X},T) = \mathbb{E}[g(\mathbf{X},T,\mathbf{U}) \mid \mathbf{X},T]. 
\end{eqnarray}
The notation means that the unmeasured confounders $\mathbf{U}$ are averaged out in the population conditional on the fixed $\mathbf{X}$ and $T$, i.e., $\mathbb{E}_\mathbf{U}[g(\mathbf{X},T,\mathbf{U})] = \mathbb{E}[g(\mathbf{X},T,\mathbf{U}) \mid \mathbf{X},T]$. 

\quad \\ \noindent \textbf{BP Model}. 

\textbf{In the first stage}, we can obtain the residual $\mathbf{V}$ from $\mathbf{V} = T - f(Z,\mathbf{X})$, and $f(Z,\mathbf{X})$ can be identified by $f(Z,\mathbf{X}) = \mathbb{E}[T \mid Z, \mathbf{X}]$:
\begin{eqnarray}
\hat{\mathbf{V}} = T - \mathbb{E}[T \mid Z, \mathbf{X}] = T - \hat{f}(Z,\mathbf{X}). 
\end{eqnarray}
where we can use machine learning methods to estimate the expectation $\mathbb{E}[T \mid Z, \mathbf{X}]$, such as kernel-based regression and neural networks regression. 

\textbf{In the second stage}, the conditional distribution of the unmeasured confounders $\mathbf{U}$ is related to $\{Z, \mathbf{X},T\}$ only through the residual $\mathbf{V}$ \cite{wooldridge2005unobserved, wooldridge2015control}:
\begin{eqnarray}
P(\mathbf{U} \mid Z,\mathbf{X},T) = P(\mathbf{U} \mid Z,\mathbf{X},\mathbf{V}) = P(\mathbf{U} \mid \mathbf{V}).
\end{eqnarray}
Then, the consistent estimator of the ASF is:
\begin{eqnarray}
& \hat{g}^{\prime}(\mathbf{X},T,\mathbf{V}) = \mathbb{E}[Y \mid \mathbf{X},T,\mathbf{V}], \\
& \text{ASF}(\mathbf{X},T) = \mathbb{E}[\hat{g}^{\prime}(\mathbf{X},T,\mathbf{V}) \mid \mathbf{X},T]  = \mathbb{E}_{\mathbf{V}}[\hat{g}^{\prime}(\mathbf{X},T,\mathbf{V})], \\
& \Hat{\text{ASF}}(\mathbf{X},T) = \frac{1}{n} \sum_{i=1}^{n}\hat{g}^{\prime}(\mathbf{X},T,\mathbf{V}).
\end{eqnarray}
where we can use machine learning methods to estimate the expectation $\hat{g}^{\prime}(\mathbf{X},T,\mathbf{V})$, such as kernel-based regression and neural networks regression. 

\subsubsection{General CFN Estimator}

Althrough CFN estimators have been widely used for solving the unmeasured confounders in causal inference, one critical limitation is that CFN usually breakdown under complex non-linear models. Besided, CFN requires that the residual obtained from the treatment outcome regression is linearly related to the unmeasured confounder, i.e., the structural treatment process assumptions, and the results is reliable only when the models are specified correctly. 

Based on the concept of variational autoencoder (VAE) \cite{kingma2013auto}, some works study the proxy variable for unmeasured confounders and try to use the proxy to reconstruct the unmeasured confounders \cite{louizos2017causal, zhang2020treatment, wu2022betaintactvae}. Motivated by this, \cite{puli2020general} develop the general control function method (GCFN) to construct general control functions and estimate effects.

With the control function that satisfies the ignorability and positivity assumptions, GCFN does not need the additive separation assumption and simplify the causal effect estimation as outcome regression on the treatment and the control function. The observation data can be sampled from: 
\begin{eqnarray}
\label{eq:GCFN_ZT}
T & = & f(Z,\mathbf{X},\mathbf{V}), \\
\label{eq:GCFN_TY}
Y & = & g(\mathbf{X},T,\mathbf{U}). 
\end{eqnarray}
Then, the control functions can be characterized:
\begin{theorem}
\textbf{Meta-identification.} The causal effect is identified by the joint distribution $q(Z,\mathbf{X},\mathbf{V},T)$ over the control function $\hat{\mathbf{V}}$ and the observables $\{Z,\mathbf{X},T\}$:
\begin{eqnarray}
\mathbb{E}_{\hat{\mathbf{V}}}[Y \mid T, \hat{\mathbf{V}}] = \mathbb{E}_{\hat{\mathbf{V}}}[Y \mid \text{do}(T), \hat{\mathbf{V}}] = \mathbb{E}[Y \mid \text{do}(T)].
\end{eqnarray}
With the following assumptions:
\begin{itemize}
  \item (A1) $\hat{\mathbf{V}}$ satisfies the reconstruction property: the treatment $T$ can be represented by $\{Z,\mathbf{X},\hat{\mathbf{V}}\}$;
  \item (A2) The IVs $Z$ are independent of control functions, confounders and residuals, i.e., $Z \perp \{\mathbf{X},\mathbf{U},\mathbf{V},\hat{\mathbf{V}}\}$;
  \item (A3) Fixing the general control function $\hat{\mathbf{V}}$, the strong IVs can set treatment to any value.
\end{itemize}
\end{theorem}
Then, the control function $\hat{\mathbf{V}}$ satisfies ignorability and positivity:
\begin{eqnarray}
q(Y \mid T,\hat{\mathbf{V}}) = q(Y \mid \text{do}(T),\hat{\mathbf{V}}), \\
q(\hat{\mathbf{V}}) > 0 \Rightarrow q(T \mid \hat{\mathbf{V}}) > 0.
\end{eqnarray}

\quad \\ \noindent \textbf{GCFN}. \\
Following \cite{guo2016control,louizos2017causal} and \cite{higgins2016beta}, GCFN's first stage called variational decoupling (VDE) constructs general control functions by using VAE and recovering the residual variation in the treatment given the IV. This yields an evidence lower bound (ELBO) of VAE to reconstruct the latent variables:
\begin{eqnarray}
& & L(\theta, \phi, \xi \mid Z,\mathbf{X},T) \nonumber \\
& = & (1+\lambda) \mathbb{E}_{q_{\theta}(\mathbf{V} \mid Z,\mathbf{X},T)}\text{log}{p_\phi}(T \mid Z, \mathbf{X}, \mathbf{V}) \nonumber \\
& - & \lambda D_{KL}(q_{\theta}(\mathbf{V} \mid Z,\mathbf{X},T) \| p_{\xi}(\mathbf{V}))
\end{eqnarray}
where $\lambda$ is the hype-parameter that is used to balance the reconstruction term and the KL term in the beta-VAE. $p_{\xi}(\mathbf{V})$ and ${p_\phi}(T \mid Z, \mathbf{X}, \mathbf{V})$ are real (posterior) probability distributions, $q_{\theta}(\mathbf{V} \mid Z,\mathbf{X},T)$ is the estimated probability distributions by neural networks with parameter $\theta$. $D_{KL}(\cdot)$ denotes the Kullback-Leibler (KL) divergence. By maximizing the above objective function, we can sample the control function $\hat{\mathbf{V}}$ from the observables $\{Z,\mathbf{X},T\}$. 

VDE provides a general control function $\hat{\mathbf{V}}$ and its marginal distribution $q_{\theta}(\mathbf{V})$. Using VDE's control function, GCFN's second stage estimates effects via regression. Other confounder adjusting/control methods like matching/balancing methods \cite{li2016matching,kuang2020causal,athey2018approximate}, doubly robust methods \cite{bang2005doubly} and representation learning methods \cite{johansson2016learning,shalit2017estimating,hassanpour2020learning,wu2022learning} can be used for outcome regression:
\begin{eqnarray}
\text{CATE} = \mathbb{E}[Y(T=t) - Y(T=0) \mid \mathbf{V}, \mathbf{X}]. 
\end{eqnarray}

Further, \cite{puli2020general} develop semi-supervised GCFN to construct general control functions using subsets of data that have both IV and confounders observed as supervision; this needs no structural treatment process assumptions. 

\subsubsection{Conditional Variational Autoencoder Estimator}

\begin{figure}
\begin{center}
\includegraphics[width=0.45\linewidth]{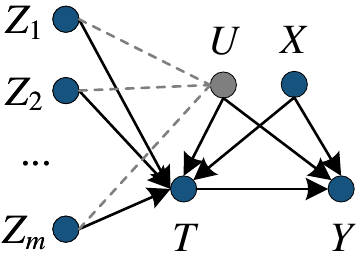}
\caption{Causal graph of Confounded IV. Dashed lines represent unknown causality. }\label{fig:cvaeiv}
\end{center} 
\end{figure} 

Due to untestable Exclusion and Independent restrictions, finding a valid IV is always a tricky problem. To relax the restriction, Wang et al. \cite{wang2022estimating} focus on estimating treatment effects with more accessible confounded instruments that violate the unconfounded instruments assumption, i.e., $\{Z_1, Z_2, \cdots, Z_m\} \not \perp \mathbf{U}$. Inspired by deep conditional variational autoencoder, they aim to generate a
substitute of unmeasured confounder that obeys strong ignorability, such that ${Y} \perp {T} \mid \mathbf{U}, \mathbf{X}$. To achieve the ignorability, CVAE-IV \cite{wang2022estimating} model a substitute $\hat{\mathbf{V}}$ based on the statistical principle $Y \perp \left\{{Z}_i\right\}_{i=1}^m \mid T, \mathbf{X}, \hat{\mathbf{V}}$, which states that
the outcome and IV candidates are conditionally independent given the
treatment, observed covariates and the generated $\hat{\mathbf{V}}$.

\quad \\ \noindent \textbf{CVAE-IV}. \\
\textbf{In the first stage}, with multiple confounded IVs $\mathbf{Z} = \left\{{Z}_i\right\}_{i=1}^m$, as shown in Fig. \ref{fig:cvaeiv}, CVAE-IV \cite{wang2022estimating} constructs a conditional variational autoencoder to generate the
confounder substitute $\hat{\mathbf{V}}$. Specifically, they apply the variational inference to model the conditional distribution $P(Y, \mathbf{Z} \mid T, \mathbf{X})$ as follow:
\begin{eqnarray}
\nonumber
\log P(Y, \mathbf{Z} \mid T, \mathbf{X}) 
\geq 
\mathbb{E}\left[\log P_\theta\left({Y}, {\mathbf{Z}} \mid T,\mathbf{X},\hat{\mathbf{V}}\right)\right] \\
- 
D_{K L}\left(Q_\phi\left(
\hat{\mathbf{V}} \mid T, Y, {\mathbf{Z}}, \mathbf{X}\right) \| P\left(\hat{\mathbf{V}} \mid T, \mathbf{X}\right)\right)
\end{eqnarray}
where $D_{K L}$ refers to the $\mathrm{KL}$-divergence between variational posterior and the underlying one, $P_\theta$ is the decoder model and $Q_\phi$ is the encoder model. By forcing the underlying posterior $P\left(\hat{\mathbf{V}} \mid T, \mathbf{X}\right)$ to follow the normal distribution. 

We use networks $f_Y$ and $f_Z$ to regress the outcome and instruments as well as minimize the evidence lower bound (ELBO) of CVAE as objective to reconstruct the latent variables $\hat{\mathbf{V}}$:
\begin{eqnarray}
\mathcal{L} &=& \mathcal{L}_{Rec} + \mathcal{L}_{Chol} + \lambda \mathcal{L}_{KL}\\
\mathcal{L}_{Rec}&=&\sum_i^n[(y_i - f_Y(t_i, \mathbf{x}_i,  \mathbf{v}_i))^2] / Var(Y), \nonumber \\
\mathcal{L}_{Chol} &=& \sum_i^n[(\mathbf{z}_i - f_Z(t_i, \mathbf{x}_i,  \mathbf{v}_i))^2], \nonumber\\
\mathcal{L}_{KL} &=& D_{K L}\left(Q_\phi\left(
\hat{\mathbf{V}} \mid T, Y, {\mathbf{Z}}, \mathbf{X}\right) \| P\left(\hat{\mathbf{V}} \mid T, \mathbf{X}\right)\right), \nonumber
\end{eqnarray}
where the $\lambda$ controls the variance of the reconstructed output.

\noindent \textbf{In the second stage}, we fit the observational outcome using two regression functions $g_{\psi_1}$ and $g_{\psi_2}$ , which are parametrized by deep networks with $\psi_1$ and $\psi_2$:
\begin{eqnarray}
\mathcal{L}_{Reg}=\sum_i^n[(y_i - g_{\psi_1}(t_i, \mathbf{x}_i)-  g_{\psi_2}(\mathbf{v}_i) )^2].
\end{eqnarray}
Then, we predict the counterfactual outcome $Y(t)$ and CATE with the trained regression model $\{\psi_1,\psi_2\}$:
\begin{eqnarray}
Y(t,\mathbf{x},\mathbf{v}) = g_{\psi_1}(t, \mathbf{x})+  g_{\psi_2}(\mathbf{v}), \\
CATE = Y(t,\mathbf{x},\mathbf{v}) - Y(0,\mathbf{x},\mathbf{v}). 
\end{eqnarray}
By constructing the CVAE-IV model to generate a ignorable confounder substitute, we
isolate the influence of the unmeasured confounder from the estimation on
conditional treatment effect.

\subsection{Limitation and Future Work}

\subsubsection{Limitation}

\noindent \textbf{Inverse Relationship}. In the structural assumption, CFN implicitly require 
a one-to-one mapping (or Inverse Relationship) between the residuals $\mathbf{V}$ from treatment regression and the unmeasured confounders $\mathbf{U}$. Otherwise, even if we recover the residuals perfectly, we cannot control the unmeasured confounders. For example, if $\mathbf{V}=sin(\mathbf{U})$, then we control for $\mathbf{V} = 1$, but $\mathbf{U}$ still has infinitely many possibilities, which we cannot discuss and analyze. 

\noindent \textbf{Invalid IV and Weak IV}.  The performance of these methods relies on the well-predefined IVs that satisfy three instruments restrictions (i.e., IV does not have a direct effect on the outcome variable, only indirectly through the treatment variable), which is untestable and leads to finding a valid IV becomes an art rather than science. Therefore, how to use invalid IV or wark IV to implement CFN is still an open problem. 

\subsubsection{Future Work}

\noindent \textbf{Variational Autoencoder}. Inverse relationship between the residuals $\mathbf{V}$ and the unmeasured confounders $\mathbf{U}$ means that we can achieve indirect control of $\mathbf{U}$ by controlling the residuals $\mathbf{V}$. So, naturally, why don't we just recover $\mathbf{U}$? Based on the concept of variational autoencoder (VAE) \cite{kingma2013auto}, some works study the proxy variable for unmeasured confounders and try to use the proxy to reconstruct the unmeasured confounders \cite{louizos2017causal, zhang2020treatment, wu2022betaintactvae}. Motivated by this, \cite{puli2020general} develop the general control function method (GCFN) to learn the distribution of unmeasured confoudners and estimate effects. 

\noindent \textbf{Confounded IV}. In reality, the acquisition of valid IV is a tricky project, so Wang et al. \cite{wang2022estimating} proposes to use confounded IV, having a direct effect on the outcome variable but indirectly through the treatment and confounders, instead of valid IV to recover unmeasured confounders. By considering the conditional independence between confounded instruments and the outcomes, CVAE-IV \cite{wang2022estimating} generates a substitute of the unmeasured confounder with a conditional variational autoencoder. Therefore, the exploration of invalid IV is a promising research line for the future.

\section{Evaluating Instrumental Variables}
\label{sec:ivtest}

\begin{figure}
\begin{center}
\includegraphics[width=0.92\linewidth]{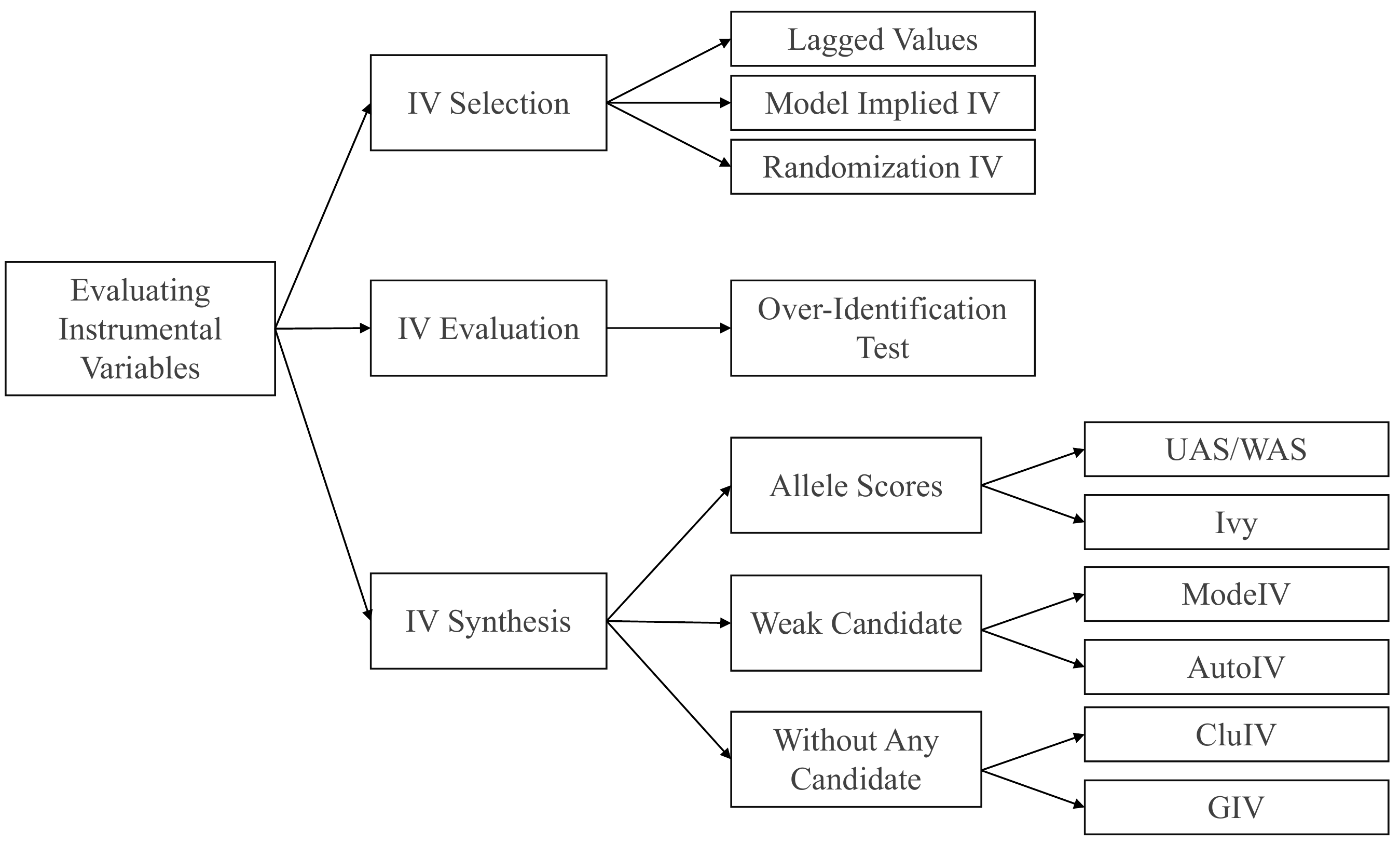}
\caption{Evaluating of Instrumental Variables.}\label{fig:evaluate}
\end{center} 
\end{figure} 

In Section \ref{sec:2SLS} \& \ref{sec:cfn}, we have introduce how to implement two-stage regression with IV for treatment effect estimation. One limitation is that, these methods require a strong and valid IV\footnote{The instrument must be correlated with the endogenous treatment variables. If this correlation is strong, then the instrument is said to have a strong first stage. A weak correlation may provide misleading inferences about parameter estimates and standard errors \cite{nichols2006weak}.} for treatment regression, which is rare in reality. In this Section, we summarize three methods for selecting IV, i.e., Lagged Values, Prior Knowledge of Causal Graph and Randomized Controlled Trials, and provide over-identification test for IV's exclusion restriction. 
Subsequently, we also introduce several machine learning algorithms for strong IV generation, i.e., Summary IVs. The overall skeleton is shown in the Fig. \ref{fig:evaluate}. 

\subsection{IV Selection}
\label{sec:test}

The above IV methods are reliable if and only if the IVs we found only affect the outcomes through its strong association with treatments. 
Finding suitable IVs still is a challenge for the IV methods \cite{bollen2012instrumental}. Next, we will introduce several methods to find or test IVs.

\textbf{Lagged Values}. With panel data, a common strategy of finding IV is to use the lagged values as IVs for the current treatments \cite{anselin1988spatial}.
For example, \cite{angrist1991does} estimated the causal
effect of compulsory schooling on earnings by using quarter of
birth as an IV for education. \cite{axinn2001mass} used characteristics of the respondent's childhood, husband's childhood, and parents and husband's parent as IVs to predict the respondent's probability to send their children to school in the future, and then used the predicted value from this model as an independent variable in the prediction of contraceptive use. 

\textbf{Model Implied Instrumental Variables}.
A second strategy draws IVs from among the observed variables is Model Implied Instrumental Variables (MIIVs), taken from \cite{bollen1996alternative,bollen2004automating,bollen2019model}. In MIIVs, a prior knowledge of causal graph is used to build  the model structure, which tells the researcher which observed variables can serve as IVs and which cannot. Closely related to the MIIV
method is the directed acyclical graph (DAG), \cite{brito2002graphical,pearl2010foundations} gave rules to select the variables that can serve as IVs: the correlation of a variable with the residual term of the outcome predict equation is zero \cite{bollen2019model}.

\textbf{Randomization Instrumental Variables}.
In instrumental variable literature, researchers usually implement Randomized Controlled Trials (RCTs) to sample a random variable as IV to intervene the received treatments, called intention-to-treat variable, such as Oregon health insurance experiment \cite{finkelstein2012oregon} and effects of military service on lifetime earnings \cite{angrist1990lifetime}, which are too expensive to be universally available.
Sometimes, there might be randomization introduced by “nature” \cite{rosenzweig2000natural}, called natural experiments, such as twin births, gender, and weather events.

\subsection{IV Evaluation}
\label{sec:eval}

Regardless of IVs selected by which prior, we must evaluate the IVs' quality: a valid IV that only affects the outcome through its strong association with treatment options, called exclusion assumption.
If the structure assumptions for IV are dissatisfied and the correlation is weak, then the instrument may provide misleading inferences about parameter estimates and standard errors \cite{nichols2006weak, bruce2022econometrics}.

\textbf{Over-Identification Test}. When the number of IVs is more than the need for just-identification, i.e., there are more IVs than the number of treatments, one can test the exogeneity of IVs. The over-identification tests construct a null hypothesis that all IVs are exogenous variables versus
the alternative hypothesis that at least one IV violates exogeneity (correlates with the residuals from the two-stage IV regression). 
In linear setting, \cite{sargan1958estimation} gave a known over-identification tests for IVs:
\begin{eqnarray}
    p = \frac{\epsilon^{\prime} \bar{Z} (\bar{Z}^{\prime}\bar{Z})^{-1}\bar{Z}^{\prime}\epsilon}{\epsilon^{\prime}\epsilon / n} \sim \mathcal{X}^2, 
\end{eqnarray}
where $\epsilon$ are the residuals from the two-stage IV regression, and $\bar{Z}$ is another instrumental variable (Over Identification) not involved in the regression of causal effects.  Asymptotically, the test statistic $p$ follows a chi square distribution and the degrees of freedom equal to the number of IVs beyond the need for just-identification \cite{wooldridge2010econometric}. 
Besides, \cite{basmann1960finite} proposed a similar over-identification tests with F-distribution. \cite{kirby200910} developed several variants for homoscedastic disturbances. Considering heteroscedastic-consistent, \cite{hansen1982large, hayashi2011econometrics} designed a test statistic for GMM-IV models.

\subsection{IV Synthesis}
\label{sec:IVsyn}

Strong and valid IVs are hardly satisfied in practice. 
Fortunately, with the advent of machine learning, researchers have found some data-driven algorithms to automatically synthesize strong IV from additional data information under some assumptions. 
Practitioners combine more commonly available IV candidates—which are not necessarily strong, or even valid, IVs—into a single “summary” that is plugged into causal effect estimators in place of an IV \cite{kuang2020ivy}.

\subsubsection{Allele Scores}
\label{sec:summary}

In Mendelian randomization (MR) \cite{burgess2015mendelian}, a growing number of works have been proposed to synthesize a summary IV by combining widely availabel IV candidates. 
\cite{burgess2016combining} shows that summary IV can be reproduced using summarized data on genetic assocaitions with the treatment and the outcome, and a representative approach that combines the IV candidates into a summary variable is unweighted/weighted allele scores \cite{burgess2017review,burgess2013use,davies2015many} (UAS/WAS). 
UAS/WAS synthesize a summary variable of genetic contribution towards elevating the risk factor, which serve as reliable IVs to infer causal effect among clinical variables, only if genetic variants associated with a risk factor are actually all independent valid IVs \cite{sebastiani2012naive,burgess2017review}. 

\quad \\ \noindent \textbf{UAS}. \\
In Mendelian randomization (MR), we can use genetic variants to as IV candidates for IV synthesis. We assume $K$ genetic variants $\mathbf{G}=\{G_1,G_2,\cdots,G_K\}$ are actually independent weak IVs, and use them as IV candidates. Then we can obtain UAS:
\begin{eqnarray}
UAS_{IV} = \frac{1}{K} \sum_{j=1}^K G_j, 
\end{eqnarray}
where $K$ denotes the number of IV candidates, and $G_j$ denotes the $j$-th IV candidate.
Factually, UAS takes the average of IV candidates.

\quad \\ \noindent \textbf{WAS}. \\
In addition to an unweighted standard allele score where each risk-increasing allele contributed the same value to the allele score, WAS weights each candidate based on the associations with the treatment: 
\begin{eqnarray}
WAS_{IV} = \frac{1}{K} \sum_{j=1}^K W_j G_j, 
\end{eqnarray}
where $W_j$ denotes the weights that are the same as the coefficients from the treatment regression stage in the 2SLS analysis. In addition, some other weight estimation methods for calculating relevance and importance can be used as an alternative. 

\quad \\ \noindent \textbf{Ivy}. \\
Allele scores require strong assumptions, i.e., all IV candidates are weak IVs for estimation. 
To relax these assumptions, \cite{kuang2020ivy} require more than half of the variables in the IV candidates are valid, and then
propose a generalized allele scores to combine valid IV candidates and invalid candidates in a robust manner, with the following steps: (1) Identify Valid IV Candidates and their Dependencies; (2) Estimate Parameters of the Candidate Model; and (3) Synthesize IV and Estimate Causal Effect.

\subsubsection{Weak Candidates}
\label{sec:generation}

Most of Allele Scores follow the assumption that IV candidates are actually all independent weak IVs, which is actually difficult to meet. In this subsection, we review some more weaker assumptions for IV Synthesis. 

\quad \\ \noindent \textbf{ModeIV}. \\
\cite{hartford2021valid} no longer requires more than half the number of valid instrumental variables in the candidate set, but proposes that each estimate in the tightest cluster of estimation points from each IV candidate is approximately causal effects and these IV candidates are valid. 
ModeIV \cite{hartford2021valid} will iterate over all the elements in the set of instrumental variable candidates $\mathbf{G}=\{G_1,G_2,\cdot,G_K\}$ and plug $G_j$ into the instrumental variable regression method to estimate the causal effects $\tau_{G_j}$. Then, the outcomes $\{\tau_{G_j}\}_{j=1}^K$ from the valid instrumental variables must all converge to the same value, and IV candidates in the tightest cluster of estimation points just are valid IVs. 

\quad \\ \noindent \textbf{AutoIV}. \\
Furthermore, AutoIV \cite{yuan2022auto} generate IV representations based on independence conditions and mutual information, with the assumption that all variables in the IV candidates $\mathbf{G}$ are independent of the unmeasured confoudners $\mathbf{U}$, i.e., $\mathbf{G} \perp \mathbf{U}$. 
Given the obaservational data $D=\{\mathbf{X}, \mathbf{G}, T, Y\}$, AutoIV \cite{yuan2022auto} learn a disentangled representation $\mathbf{Z} = \phi(\mathbf{G})$ based on independence conditions: 
\begin{eqnarray}
& & \hat{\phi} = \arg \min_{\phi} (T - f(\phi(\mathbf{G}), \mathbf{X}))^2, \\
& \text{s.t.  } & \phi(\mathbf{G}) \perp \mathbf{X}, \nonumber \\
& &  \phi(\mathbf{G}) \perp Y \mid T, \mathbf{X}, \nonumber
\end{eqnarray}
where $f(\cdot)$ denotes a regression network of $\phi(\mathbf{G}), \mathbf{X}$ to predict treatment variables. According to the independence conditions, AutoIV \cite{yuan2022auto} obtain valid IVs that does not have a direct effect on the outcome variable, only indirectly through the treatment variable. 

To learn relevance and exclusion, AutoIV \cite{yuan2022auto} construct a mutual information estimation network to optimize the network. Take two any random variables $X$ and $Y$ as an example, the log-likelihood loss function of variational approximation $Q_{\theta_{XY}}(Y|\phi_X(X))$ with $n$ samples is given as:
\begin{equation}\label{equ-zx-lld}
    \mathcal{L}_{XY}^{LLD}=-\frac{1}{n}\sum_{i=1}^n{\log{Q_{\theta_{XY}}(y_i|\phi_X({x}_i))}}.
\end{equation}

They minimize Eq. (\ref{equ-zx-lld}) to get optimal variational approximation $Q_{\hat{\theta}_{XY}}(Y|\phi_X(X))$ with parameters $\hat{\theta}_{XY}$. 
To increase the relevance between the IV representations and the treatment, they maximize the mutual information between them:
\begin{equation}\label{equ-zx-mi}
    \begin{aligned}
        \mathcal{L}_{XY}^{MI}=\frac{1}{n^2}\sum_{i=1}^n\sum_{j=1}^n(\log{Q_{\theta_{XY}}(y_{i}|\phi_X(x_i))}-\\
        \log{Q_{\theta_{XY}}(y_{j}|\phi_X(x_i))}),
    \end{aligned}
\end{equation}

Besides, they also model the conditional mutual information $\mathcal{L}_{XY\mid V}^{MI}$ conditional on random variable $Z$ as:
\begin{equation}\label{equ-zx-cmi}
    \begin{aligned}
        \mathcal{L}_{XY\mid V}^{MI}=\frac{1}{n^2}\sum_{i=1}^n\sum_{j=1}^n(\omega_{i j}(\log{Q_{\theta_{XY}}(y_{i}|\phi_X(x_i))}-\\
        \log{Q_{\theta_{XY}}(y_{j}|\phi_X(x_i))})),
    \end{aligned}
\end{equation}
where, $\omega_{ij}={\rm{softmax}}(e^{-\frac{{\Vert \boldsymbol{x}_{i}-\boldsymbol{x}_{j} \Vert}^2}{2\sigma ^2}})$ is the conditional weight of each pair of positive and negative samples.  

Based on the $\mathcal{L}_{XY}^{LLD}$ and $\mathcal{L}_{XY\mid V}^{MI}$ operators, AutoIV \cite{yuan2022auto} (1) maximize $\mathcal{L}_{\mathbf{G}T}^{MI}$ to optimize the IV representations $\phi(\mathbf{G})$ for relevance condition; (2) minimize $\mathcal{L}_{\mathbf{G}Y \mid T}^{MI}$ to optimize the IV representations $\phi(\mathbf{G})$ for exclusion condition; and (3) minimize $\mathcal{L}_{\mathbf{G}\mathbf{X}}^{MI}$ to optimize the IV representations $\phi(\mathbf{G})$ for observed confounders independence condition.

\subsubsection{Without Any Candidates}
\label{sec:giv}

\textbf{Limitation}. 
Although the above IV generation methods no longer require manually selected pre-defined IVs selected, they all require a high-quality IV candidates' set with at least half valid IVs or unconfounded IV assumption, which is unrealistic in practice due to cost issues and lack of expert knowledge. These methods still cannot get rid of the dependence on predefined candidate sets. Therefore, it is highly demanded to model IVs and implement a data-driven approach to automatically obtain valid IVs directly from the observed variables $\{\mathbf{X},T,Y\}$. 

In 2021, the idea of using clustering methods to generate instrumental variables started to present, such as CluIV \cite{sokolovska2021role} and GIV \cite{wu2022treatment}.
Under a more practical setting without any candidates, GIV \cite{wu2022treatment} proposes a novel algorithm (Meta-EM) to model latent GIV and implement a data-driven approach to automatically reconstruct valid Group IVs directly from the observed variables, beyond hand-made IV candidates. 

\begin{figure}[t]
\centering
\includegraphics[width=0.9\columnwidth]{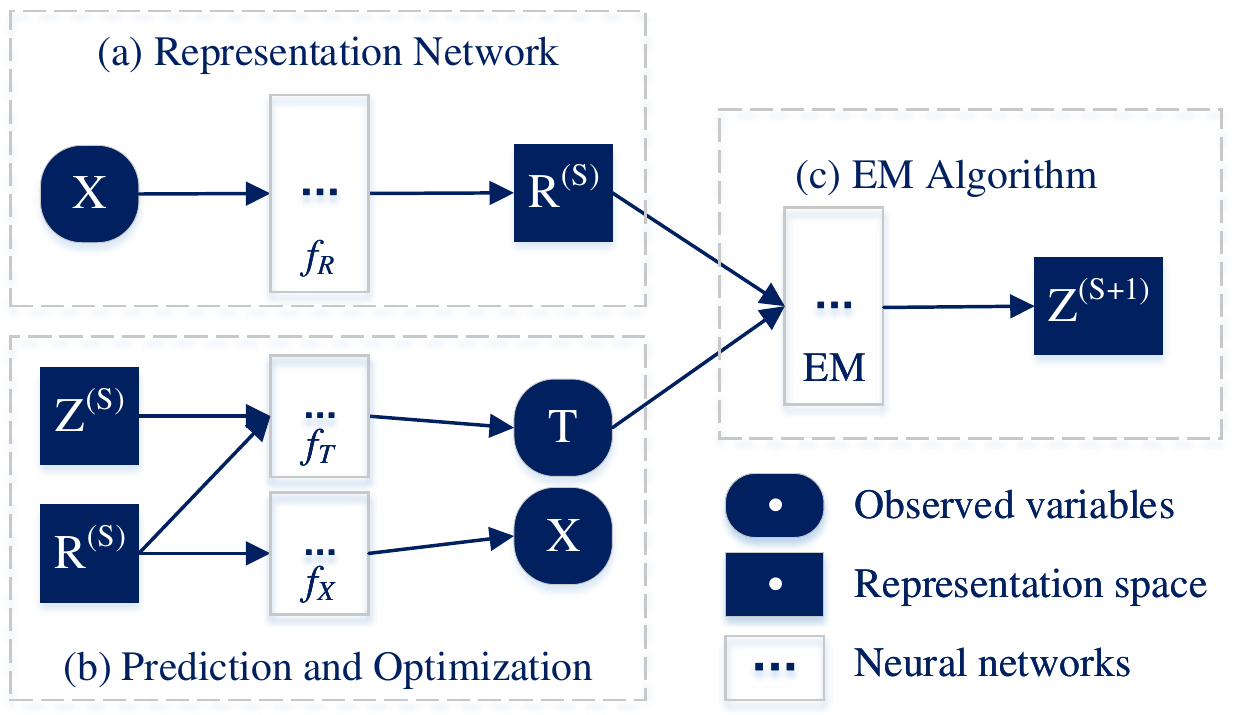} 
\caption{Overview of Meta-EM Architecture.}
\label{fig15}
\end{figure}

\quad \\ \noindent \textbf{GIV}. \\
With the advent of the big data era, a variety of observation databases collected from different sources have been established, which may contain the same treatment effect mechanism (from treatment to outcome) but different treatment assignment mechanisms (from covariates to treatment). Here, the omitted source label can serve as a latent multi-valued IV, which only affects the outcome through its strong association with offer decisions. 

Therefore, as shown in Fig. \ref{fig15}, Wu et al. \cite{wu2022treatment} propose a non-linear Meta-EM to (1) map the raw data into a representation space to construct Linear Mixed Models for the assigned treatment variable; (2) estimate the distribution differences and model the GIV for the different treatment assignment mechanisms; and (3) adopt an alternating training strategy to iteratively optimize the representations and the joint distribution to model GIV for IV regression. Empirical results demonstrate the advantages of our Meta-EM compared with state-of-the-art methods.


\section{Available Datasets and Codes/Packages}
\label{sec:code}

\subsection{Datasets}

In real-world applications, it's thorny to find a strictly valid instrumental variable from observational data, due to the untestable exclusion and unconfounded conditions. In short, the predefined IVs and IV candidates selected by human effort might be invalid IVs that do not strictly satisfy the conditions of the valid IVs, without enough prior knowledge for valid IVs. Besides, in observational dataset, the
ground truth dose-response function (ATE, ATT, CATE or ITE) is not available, due to the lack of the counterfactual outcome. Hence, the datasets used in the IV-based works are often (semi-)synthetic datasets, such as Demand \cite{hartford2017deepiv} and Toy Datasets \cite{bennett2019deep,lewis2020agmm}. Some datasets combine the prior specific knowledge and the observational control dataset together to create the datasets. We detail the available benchmark datasets, as follows:

\textbf{Low-dimensional Toy \cite{bennett2019deep,lewis2020agmm}}.
In low-dimensional cases, \cite{bennett2019deep} generated data via the following process:
\begin{eqnarray}
& Y = g(T) + U + \delta, T = Z + U + \gamma. \\
& Z \sim \text{Uniform}(-3,3), U \sim \mathcal{N}(0,1), \delta,\gamma \sim \mathcal{N}(0,0.1). \nonumber
\end{eqnarray}
Similarity, \cite{lewis2020agmm} consider the following data generating processes:
\begin{eqnarray}
& Y = g(T) + U + \delta, T = \gamma Z + (1-\gamma)U + \gamma. \\
& Z \sim \mathcal{N}(0,2), U \sim \mathcal{N}(0,2), \delta,\gamma \sim \mathcal{N}(0,0.1). \nonumber
\end{eqnarray}
Keeping the data generating process fixed, \cite{bennett2019deep,lewis2020agmm} design various true response function $g$ between the following cases:
\begin{eqnarray}
\textbf{sin: } & g(T) = sin(x), &   \textbf{step: } g(T) = 0,  \nonumber \\
\textbf{abs: } & g(T) = |x|,  &  \textbf{linear: } g(T) = x. \nonumber
\end{eqnarray}

\textbf{MNIST \cite{bennett2019deep}}.
Similar to \cite{hartford2017deepiv}, in high-dimensional cases, \cite{bennett2019deep} use same data generating process introduced in Low-dimensional Toy, based on the MNIST dataset \cite{lecun1998gradient}, but replace $T$ and $Z$ with MNIST images: 
\begin{eqnarray}
& T := \text{RandomImage}(\pi(T)), 
Z := \text{RandomImage}(\pi(Z)). \nonumber
\end{eqnarray}
where $\pi(t) = \text{round}(\text{min}(\text{max}(1.5t+5,0),9))$ is a transformation function that maps input $t$ to an integer range from 0 to 9, and the RandomImage($d$) is a function that samples a image from the digit label $d$. The images are $28 \times 28 = 784$-dimensional digit matrices. 

\textbf{Demand \cite{hartford2017deepiv}}. The demand simulation design is from \cite{hartford2017deepiv}, which describes an airline scenario. In this simulation, the airline wants to estimate the effect of prices $T$ (i.e., treatment) on passenger ticket sales $Y$ (i.e., outcome). We assume that the fuel price $Z$, the customer types $X_1$, the time of year $X_2$, and the conferences $U$ are the pre-treatment variables $V$, where the instrumental varialble is $Z$, the observable confounders are $X=\{X_1,X_2\}$ and the unmeasured confounder is $U$. The simulation data is generated by:
\begin{eqnarray}
T & = & 25 + (Z + 3) \psi(X_2) + U, \\
Y & = & 100 + (10 + T)X_1 \psi(X_2) - 2T + \epsilon, \\
\psi(X_2) & = & \Scale[0.95]{ 2 \left( \frac{1}{600} (X_2 - 5)^4 + \text{exp}[-4 (X_2 - 5)^2] + \frac{X_2}{10} - 2 \right)}, \nonumber \\
X_1 & \in & \{1,\cdots, 7\}, \quad X_2 \sim \text{unif}(0,10), \nonumber \\
Z, U & \sim & \mathcal{N}(0,1), \qquad \epsilon \sim \mathcal{N}(\rho U, 1 - \rho^2). \nonumber 
\end{eqnarray}
where, the simulation generates the latent errors $\epsilon$ with a parameter $\rho$ that is used to smoothly vary the unmeasured confounding bias in causal model. 

The target dose-response function, i.e., counterfactual function is $g(T,X) = (10 + T)X_1 \psi(X_2) - 2T$. 

\textbf{IHDP\footnote{{http://www.fredjo.com}} \cite{johansson2016learning,shalit2017estimating}} .
The Infant Health and Development Program (IHDP), from a Randomized Controlled Trial (RCT), assesses whether the future cognitive of of premature infants is affected by specialist home visits. To reduce the randomness and create a observational data, \cite{hill2011bayesian} removed a non-random subset of the treated group to induce selection bias. The dataset comprises 747 units (139 treated, 608 control) with 25 pre-treatment variables related to the children and their mothers. The treatment is the specialist home visits and the outcome is the cognitive test scores in the future. To develop instrument variables, \cite{wu2022treatment} generate 2-dimension random variables for each unit. Then, \cite{wu2022treatment} select a subset of pre-treatment variables as the confounders unobserved confounders $U$. With known treated and control potential outcome (accessible in IHDP), \cite{wu2022treatment} designs the treatment assignment policy as:
\begin{eqnarray}
\Scale[0.9]{P(T \mid Z,X) = \frac{1}{1+\exp{\left(-(\sum_{i=1}^{2}Z_i+\sum_{i=1}^{m_X}X_i)+\sum_{i=1}^{m_U}U_i)\right)}}}, \\
T \sim Bernoulli(P(T \mid Z,X)), Z_1, Z_2 \sim \mathcal{N}(0,1)
\end{eqnarray}
where ${m_X}$ and ${m_U}$ are the dimensions of $X$ and $U$ selected from the IHDP. 

\textbf{PISA \cite{pokropek2016introduction}}.
The PISA survey aims to evaluate the students’ ability to apply their knowledge and skills to real-life situations \cite{pokropek2016introduction}, covering three main domains: reading (131 items), mathematics (35 items), and science (53 items). 
\cite{schulz2011iccs,domanski2012school} selected 4951 participants in March 2009, 4041 participants in October 2009 and 3989 participants in April 2010 and there are 3472 students participated in all three rounds.  The distance to school was expressed in the number of minutes is an instrument. 
Gender and type of school (General comprehensive, Vocational with comprehensive program, and Basic vocational school) are used as covariates. 

\textbf{ALSPAC\footnote{http://www.alspac.bris.ac.uk} \cite{timpson2009does,palmer2012using}}.
The Avon Longitudinal Study of Parents and Children (ALSPAC) is a longitudinal, population-based birth cohort study from 14541 pregnant women resident in Avon, UK, with expected dates of delivery range from April 1991 to December 1992 \cite{golding2001alspac}. Similar to \cite{timpson2009does}, through selection, \cite{palmer2012using} used four adiposity-associated genetic variants as IVs for estimating the effect of fat mass on kid's bone density, based on 5509 birth cohorts.

\textbf{MR-base\footnote{https://www.mrbase.org/} \cite{hemani2018mr}}.
\cite{hemani2018mr} developed a MR-Base platform that integrates a curated database of complete GWAS results, which used genetic variants as instrumental variables. The database comprises 11 billion single nucleotide polymorphism-trait associations from 1673 GWAS and is under updated. 

\begin{table*}[ht]
\caption{Available Codes of Methods for Instrumental Variables and Causal Inference.}
\label{tab:code}
\begin{center}
\scalebox{1.0}{\begin{tabular}{ccc}
\toprule
\multicolumn{3}{c}{IV-based Methods} \\
\hline
\textbf{Method} & \textbf{Language} & \textbf{Link} \\
\hline
DeepIV & python & https://github.com/jhartford/DeepIV \\
KernelIV & matlab &  https://github.com/r4hu1-5in9h/KIV \\
DualIV & matlab & https://github.com/krikamol/DualIV-NeurIPS2020 \\
DFIV & python & https://github.com/liyuan9988/DeepFeatureIV \\
DeepGMM & python & https://github.com/CausalML/DeepGMM \\
AGMM & python & https://github.com/microsoft/AdversarialGMM \\
CBIV & python & https://github.com/anpwu/CB-IV  \\
\hline
AutoIV & python & https://github.com/junkunyuan/AutoIV \\
\hline
econML & python & https://github.com/microsoft/EconML \\
CausalDCD & python & https://github.com/anpwu/Awesome-Instrumental-Variable \\
\bottomrule
\end{tabular}}
\end{center}
\end{table*}

\subsection{Codes/Packages}
In this part, we summarize the available codes for instrumental variables and causal inference, see Table \ref{tab:code}. Besides, we merge these codes into a tool-box \textbf{CausalDCD}.

\section{Applications}
\label{sec:application}

In practical, unmeasured confounder is a common setting. Therefore, in the presence of unmeasured confounders, IV regression algorithms have a variety of applications in real-world scenarios. 

\subsection{Mendelian Randomization}

According to the Mendel’s First and Second Laws of Inheritance, when applied to independent heritable units, genotype is independent of unmeasured confounders. 
Therefore, Mendelian randomization (MR) analysis (first used by \cite{gray1991avoid}), using genetic variants as instrumental variables to estimate causal effects in the presence of unmeasured confounders \cite{youngman2000plasma,davey2003mendelian,thomas2004commentary}, is receiving increasing attention from economists, statisticians, epidemiologists and social scientists are focus \cite{smith2007capitalizing,thanassoulis2009mendelian,palmer2012using}.
The growing availability in genome-wide association studies (GWAS) facilitated discovery of genetic variants, that only affects the outcomes through its strong association with treatment factors of interest \cite{palmer2012using, von2016genetic}. 

By comparing outcomes in patients with and without human leukocyte antigen (HLA)-compatible siblings, \cite{gray1991avoid} first proposed 'Mendelian randomization' method to explore the effect of allogenic sibling bone marrow transplantation on the treatment of acute myeloid leukaemia (AML).
Mendelian randomization provides one method for assessing the causal nature of some treatment exposures \cite{davey2003mendelian}. 
\cite{timpson2009does} used two independent genetic markers (FTO and MC4R genes) of obesity as IVs and found a positive effect of fat mass on bone mineral density (BMD), i.e., higher fat mass caused increased accrual of bone mass in childhood.
\cite{palmer2012using} used multiple genetic variants as instrumental variables  for increasing statistical precision of IV estimates and for testing underlying IV assumptions. 

Use of Mendelian randomisation is growing rapidly \cite{timpson2005c, palmer2012using}. Recently, MR has been used successfully across a wide range of domains, i.e., drug target validation, drug target repurposing, side effect identification, and interpretation of high-dimensional omics studies \cite{zheng2017recent, davies2018reading}. \cite{zheng2017recent} reviewed recent developments in Mendelian randomization Studies and detailed the extensions to the basic MR design: 
including two-sample Mendelian randomization \cite{burgess2015mendelian,burgess2015using,hartwig2016two}, 
bidirectional Mendelian randomization \cite{timpson2011c}, 
two-step Mendelian randomization \cite{burgess2015network}, 
multivariable Mendelian randomization \cite{burgess2015multivariable,kemp2016using} 
and factorial Mendelian randomization \cite{ference2017association}. 
In all, MR is a flexible and robust statistical method, which uses genetic variants as IVs to identify the causal relationships from observational studies.

\subsection{Sociology and Social Sciences}

In sociology and social sciences, the purpose of causal inference is to examine the association between social network and behaviors, also known as peer effects, social contagion or induction \cite{jencks1990social,foster1997instrumental,o2014estimating}.
The peer effect means that the behavior, traits, or characteristics of an individual's peers (those he is connected to or alters) would affect his behavior\cite{o2014estimating}.
Due to contextual confounding, peer selection, simultaneity bias and measurement error, \cite{an2015instrumental} points that it is very difficult to estimate the peer effects from observational data but instrumental variables (IVs) can help to address these problems. 

Taking the city-level characteristics serve as instruments, \cite{foster1996illustration} study the effect of the neighborhood dropout rate on the individual's chance of finishing high school. To explore whether moving to a lower dropout rate would lower ones' chance of dropping out, \cite{foster1997instrumental} used characteristics of the local labor market (or city) as instruments and the results suggested that neighborhood conditions do influence an individual’s likelihood of finishing high school. 

Besides, researchers and data scientists have an increasing interest on the social network services in the Facebook, Twitter, Wechat and etc \cite{han2020social}, which are collectively called 'social media'. Adopting an instrumental variables approach, \cite{han2020social} explored the effect of social network services on social capital. \cite{han2020social} suggested that high intensity users are higher in network social capital than non-users of social network services.

\subsection{Reinforcement Learning}

In reinforcement learning (RL), an agent would take actions in an environment in order to maximize the cumulative reward \cite{minsky1954theory,watkins1989learning}. Many concepts of reinforcement learning can be found in causal inference: the treatment is the action taken by agents, the environment can be viewed as the confounder and the cumulative reward is the outcome in causal inference \cite{forney2017counterfactual,gershman2017reinforcement}. 

For reinforcement learners, the environment information is usually accessible, due to the Markov property \cite{kaelbling1996reinforcement,lei2021deep}, which satisfies the unconfoundedness assumption \cite{kallus2020confounding}. 
To obtain an unbiased reward estimation, importance sampling weighting and doubly robust policy evaluation \cite{precup2000eligibility,dudik2011doubly} are common methods adopted in RL. Under unconfoundedness assumption, there are a substantial number of variants can estimate the state-action value (Q-function) \cite{swaminathan2015counterfactual,swaminathan2017off,zou2019focused,kumor2021sequential}.

To relax the unconfoundedness assumption, \cite{xu2002efficient,chen2021instrumental,liao2021instrumental,li2021causal} introduced instrumental variables to optimize the policy for maximizing the reward. 
In the context of offline policy evaluation (OPE), \cite{chen2021instrumental} proposed improved Q-function estimators with different IV techniques and obtain competitive new techniques in recovering previously proposed OPE methods.
Using IVs, \cite{liao2021instrumental} derived a conditional moment restriction (CMR) and propose a IV-aided Value Iteration (IVVI) algorithm based on a primal-dual reformulation of CMR. In addition, \cite{li2021causal} developed a new techniques to apply IV Regression to correct for the bias in RL algorithm in the presence of time-dependency noise. 

\subsection{Recommendation System}

Another application, highly correlated with the treatment effect estimation, is recommendation system \cite{schnabel2016recommendations,swaminathan2017off,wang2019doubly,lada2019observational,yao2021survey}. Exposing the user to an item can be viewed as a specific treatment and the user's behaviour (click or activity) is the corresponding outcome. 
To elimate the bias form the unmeasured confounders and the self-selection of the users, \cite{sharma2015estimating} proposed an instrumental variable estimate of the click-through rate, where the shock is the instrument, the treatment is exposure to the focal product, and the outcome is click-through to the recommended product.
Jointly considering users’ behaviors in search scenarios and recommendation scenarios, \cite{si2022model} embedded users’ search
behaviors as instrumental variables (IVs) and implemented a two-stage regression for an unbiased estimate of causal effect.


\subsection{Computer Vision}

Computer Vision is a typical field of artificial intelligence (AI), suffering from unstable learning and lacking of generalization ability \cite{scholkopf2021towards}. To achieve a proactive defense against adversarial examples, \cite{tang2021adversarial} proposed to use the instrumental variable that achieves causal intervention. 
Using “retinotopic sampling” as IV \cite{arcaro2009retinotopic}, 
Causal intervention by instrumental Variable (CiiV) \cite{tang2021adversarial} algorithm 
implements a spatial data augmentation using different retinotopic sampling masks and learns features linearly responding to spatial interpolations. 
In Domian Adaptation, \cite{yuan2021learning} claimed that the input features of one domain are valid instrumental variables for other domains.  Inspired by this finding, we design a simple yet effective framework to learn the Domain-invariant Relationship with Instrumental VariablE (DRIVE) via a two-stage IV method.

\section{Conclusion}
\label{sec:conclusion}

\subsection{Future Direction}

Instrumental Variable has been an attractive research topic for a long time as it provides an effective way to uncover causal relationships in real-world problems. In this section, we point several lines for further research.

\subsubsection{How to find a valid IV?}
The exclusion restriction is the most critical and typically most controversial assumption underlying instrumental variables methods and we don't have any means to test it. In traditional literature, researchers implement randomized controlled trials (RCTs) to sample a random variable as IV to intervene the received treatments, which are too costly to be universally available. 
Therefore, it’s highly demanding to develop a data-driven approach to automatically obtain valid IVs. 
Fortunately, machine learning and Bayesian learning provide tools for modeling latent variables. Based on conditional independence test, it is likely to disentangle instrumental variables from these hidden variables. 
In addition, causal discovery algorithms are also a promising direction to help us automatically find instrumental variables from observed covariates.

\subsubsection{How to relax the IV assumptions? }
An instrument meets the following three assumptions: relevance assumption, exclusion assumption and unconfoundedness assumption. For exclusion assumption, we can use some mediators to block out the direct effect of IV on the outcomes to relax it. For confounded IV, we can also try to recover the unmeasured confounders affecting IV based on conditional independence constraints, and adjust it. 

\subsubsection{How to combine IV Regression with Confounder Control? }
In traditional instrumental variable regression methods, researchers always ignore the bias caused by the observed confounding variables. Even by CFN, the investigators did not control for confounding of the recovered residuals. Considering confounder balance, a more robust instrumental variable regression method is a promising direction. 

\subsubsection{How to reduce unmeasured confounding without IV? }
However, in real life, instrumental variables may not always exist, which is the norm. In the past, we have always considered observational datasets or randomized controlled experiments separately. But in fact, even if randomized controlled experiments are expensive, we can still conduct small-scale randomized controlled experiments. Considering small intervention data and a large amount of observational data, i.e., data fusion, it is possible to establish causality 
without confounding bias.

\subsection{Conclusion}

In this survey, we provide a comprehensive review of the connection between the instrumental variable methods and machine learning models. 
Combined with machine learning, we mainly introduce two typical types of methods to estimate the average treatment effect: two-stage least squares (vanilla 2SLS estimator for linear models and machine learning estimator for non-linear models) and the traditional control function method (linear estimator and non-linear estimator).
As IV-based framework relies on one structural assumption and three restrictions for identification of causal effects, 
we also review the traditional identifiability assumptions that apply in various scenarios and how to find or generate a valid IV towards these restrictions. 
The available benchmark datasets and open-source codes of those methods are also listed.
Finally, some representative real-world applications of causal inference are introduced, such as advertising,
recommendation, medicine, and reinforcement learning.

\balance
\bibliographystyle{IEEEtran}
\bibliography{ref.bib}

\end{document}